\documentclass[10pt,twocolumn,letterpaper]{article}
\usepackage{booktabs}

\usepackage{multirow}
\usepackage{makecell}
\usepackage{amssymb}
\usepackage{graphicx}
\usepackage{caption}
\usepackage{subcaption}
\usepackage{array}
\usepackage{bbding}
\usepackage{xcolor}
\definecolor{citecolor}{HTML}{0071BC}
\definecolor{linkcolor}{HTML}{ED1C24}

\newlength\savewidth\newcommand\shline{\noalign{\global\savewidth\arrayrulewidth
  \global\arrayrulewidth 1pt}\hline\noalign{\global\arrayrulewidth\savewidth}}
\newcommand{\tablestyle}[2]{\setlength{\tabcolsep}{#1}\renewcommand{\arraystretch}{#2}\centering\footnotesize}
\renewcommand{\paragraph}[1]{\vspace{1.25mm}\noindent\textbf{#1}}

\newcolumntype{x}[1]{>{\centering\arraybackslash}p{#1pt}}
\newcolumntype{y}[1]{>{\raggedright\arraybackslash}p{#1pt}}
\newcolumntype{z}[1]{>{\raggedleft\arraybackslash}p{#1pt}}

\newcommand{\app}{\raise.17ex\hbox{$\scriptstyle\sim$}}

\definecolor{deemph}{gray}{0.6}

\definecolor{baselinecolor}{gray}{.9}

\usepackage{float}

\usepackage[linesnumbered,ruled,vlined]{algorithm2e}
\usepackage{wrapfig}

\SetKwComment{Comment}{\color{green!50!black}\# }{}

\newcommand{\var}{\texttt}

\SetKwProg{Function}{def}{:}{}

\SetKwProg{For}{for}{:}{}
\SetKwProg{If}{if}{:}{}
\usepackage[linesnumbered,ruled,vlined]{algorithm2e}
\usepackage{etoolbox}
\makeatletter
\AfterEndEnvironment{algorithm}{\let\@algcomment\relax}
\AtEndEnvironment{algorithm}{\kern2pt\hrule\relax\vskip3pt\@algcomment}
\let\@algcomment\relax
\newcommand\algcomment[1]{\def\@algcomment{\footnotesize#1}}
\renewcommand\fs@ruled{\def\@fs@cfont{\bfseries}\let\@fs@capt\floatc@ruled
  \def\@fs@pre{\hrule height.8pt depth0pt \kern2pt}%
  \def\@fs@post{}%
  \def\@fs@mid{\kern2pt\hrule\kern2pt}%
  \let\@fs@iftopcapt\iftrue}
\makeatother
\usepackage{cvpr}              % To produce the CAMERA-READY version
\definecolor{golden}{RGB}{171, 107, 35}
\definecolor{cvprblue}{rgb}{0.21,0.49,0.74}
\usepackage[pagebackref,breaklinks,colorlinks,allcolors=cvprblue]{hyperref}
\usepackage{hypcap}
\def\paperID{xxx} % *** Enter the Paper ID here
\def\confName{CVPR}
\def\confYear{2024}

\title{\includegraphics[width=0.07\linewidth]{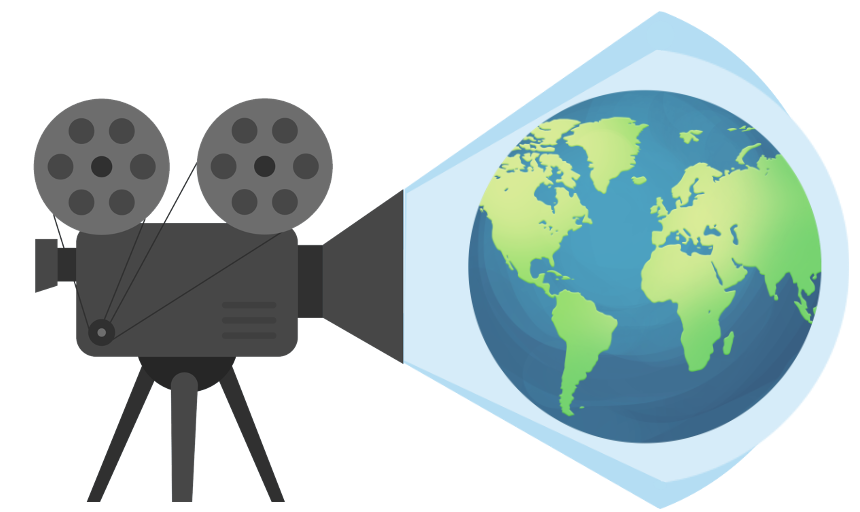} VideoWorld: Exploring  Knowledge Learning from Unlabeled Videos}
\author{Zhongwei Ren$^{1,2}$,~~Yunchao Wei$^{1}$\textsuperscript{\textnormal{$\dagger$}},~~Xun Guo$^{2}$,\\~~Yao Zhao$^{1}$,~~Bingyi Kang$^{2}$,~~Jiashi Feng$^{2}$,~~Xiaojie Jin$^{2}$\textsuperscript{\textnormal{$\dagger$}\textnormal{$\ddagger$}}\\
{\fontsize{10pt}{12pt}\selectfont $^{1}$Beijing Jiaotong University }  {\fontsize{10pt}{12pt}\selectfont$^{2}$ByteDance Seed }\\
 {\hypersetup{urlcolor=golden}
   \fontsize{10pt}{12pt}\selectfont \href{https://maverickren.github.io/VideoWorld.github.io/}{https://VideoWorld.github.io/}}
}

\usepackage{fancyhdr} 
\begin{document}

\maketitle
\thispagestyle{fancy}
\fancyhead[L]{\includegraphics[height=15pt]{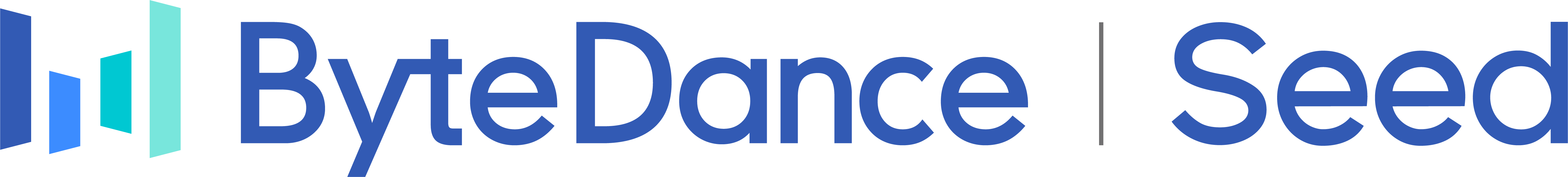}} % 在左侧添加logo
\fancyhead[R]{}
\begin{abstract}
This work explores whether a deep generative model can learn complex knowledge solely from visual input, in contrast to the prevalent focus on text-based models like large language models (LLMs). We develop \emph{VideoWorld}, an auto-regressive video generation model trained on unlabeled video data, and test its knowledge acquisition abilities in video-based Go and robotic control tasks. Our experiments reveal two key findings: (1) video-only training provides sufficient information for learning knowledge, including rules, reasoning and planning capabilities, and (2) the representation of visual change is crucial for
knowledge acquisition.
To improve both the efficiency and efficacy of this process, we introduce the Latent Dynamics Model (LDM) as a key component of VideoWorld. Remarkably, VideoWorld reaches a 5-dan professional level in the Video-GoBench with just a 300-million-parameter model, without relying on search algorithms or reward mechanisms typical in reinforcement learning. In robotic tasks, VideoWorld effectively learns diverse control operations and generalizes across environments, approaching the performance of oracle models in CALVIN and RLBench. This study opens new avenues for knowledge acquisition from visual data, with all code, data, and models open-sourced for further research.
\footnotetext{Work done when Zhongwei and Xun interned at ByteDance Seed. $^\dagger$Correspondence to Xiaojie Jin $<$\url{jinxiaojie@bytedance.com}$>$ and Yunchao Wei $<$\url{yunchao.wei@bjtu.edu.cn}$>$. $\ddagger$ Project lead.}
\end{abstract}    
\vspace{-1em}
\section{Introduction}
\label{sec:intro}

The next token prediction training paradigm has endowed large language models (LLMs)~\cite{gpt, touvron2023llama, chowdhery2023palm, bai2023qwen, anil2023palm2, yang2024qwen2, jiang2023mistral, wang2024world, gu2023text, gu2024context, jiao2024collaborativevisiontextrepresentationoptimizing} with remarkable world knowledge and intelligence, enabling them to help address complex tasks that require reasoning~\cite{wei2022cot, kojima2022large, yuan2024advancing, havrilla2024teaching, ren2024pixellmpixelreasoninglarge, han2024roserevolutionizingopensetdense, qu2023riobenchmarkreasoningintentionoriented, qu2024chatvtgvideotemporalgrounding}, planning ahead~\cite{huang2022languageplanner, song2023llm, shi2024enhancing, wang2023describe}, and decision-making~\cite{liu2024dellma, chen-etal-2024-efficient, eigner2024determinants, chen2024alignment}. 
However, language alone cannot fully capture all forms of knowledge or encompass the vast information present in the real world. In nature, biological organisms acquire knowledge primarily through visual information, rather than relying solely on language. For instance, gorillas and other primates learn vital skills like foraging and social interactions mainly through visual observation, mimicking adult behaviors without relying on language~\cite{hoppitt2013social,byrne1998learning,whiten2005conformity}.

Most existing research has focused on learning knowledge from texts or labels~\cite{team2023gemini, achiam2023gpt}, with relatively little attention to learning from pure visual signals. Some studies, such as~\cite{du2023unipi}, have explored using video data to train models for robot manipulation, but they still rely heavily on language instructions. Moreover, these tasks are often limited to single commands, without requiring complex reasoning or planning. This raises an important question: \textbf{\textit{can an AI model learn knowledge\footnote{Following prior works in AI and knowledge representation~\cite{russell2016artificial,brachman2004knowledge} which view `knowledge' as extending beyond factual information, we use `\textit{knowledge}' to broadly refer to a model's learned rules, reasoning, and planning abilities necessary for task completion. For better readability and clarity, these terms are used interchangeably in specific contexts.}
solely from visual input, akin to how a gorilla learns from its environment}?}

\begin{figure}[t]
\includegraphics[width=\linewidth]{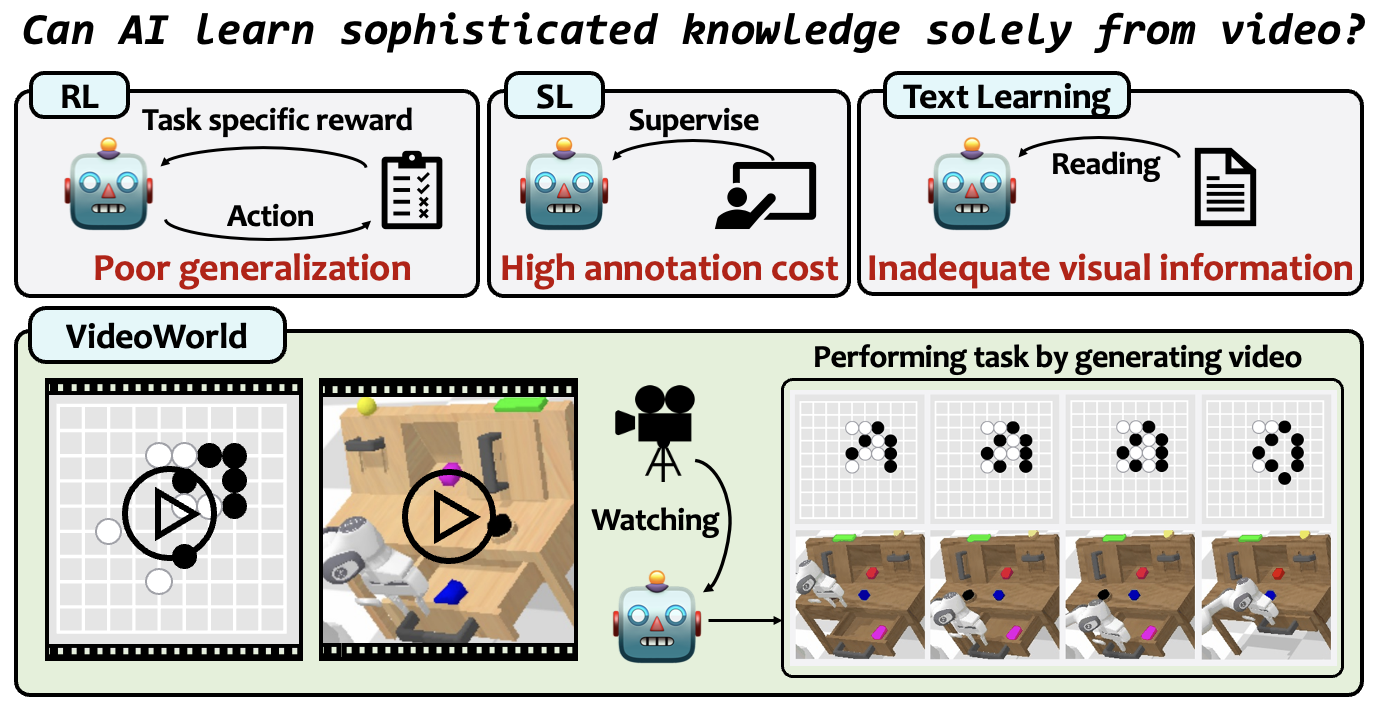}
\centering
\vspace{-2em}
\caption{VideoWorld explores learning knowledge from unlabeled videos, ranging from task-specific rules to high-level reasoning and planning capabilities. Compared to other learning methods: reinforcement learning (RL), supervised learning (SL) and text-based learning, it offers three advantages: 1) better generalization with unified visual representation for various tasks and interfaces, 2) lower manual annotation burden, and 3) learning richer real-world information than text description.
}
\label{fig:fig1_show}
% \vspace{-2em}
\end{figure}

In this work, we take an initial step toward exploring knowledge learning from raw video data by leveraging the next token prediction paradigm. To this end, we construct two experiment environments: Go and robotic manipulation (as shown in Fig.~\ref{fig:fig1_show}) to collect purely visual training data. The Go game serves as an ideal testbed for assessing a model’s ability to learn rules, reasoning, and planning: it not only has well-defined rules but also demands complex reasoning about the current state to determine the best move and forward planning to outmaneuver an opponent. Moreover, it disentangles low-level details (e.g., appearance and texture) from higher-level knowledge, making it particularly suitable for our investigation.
In parallel, we also assess the model's capability to understand robotic manipulation rules and planning, trained solely on raw video data, for control tasks in the CALVIN~\cite{mees2022calvin} and RLBench~\cite{james2019rlbenchrobotlearningbenchmark} benchmarks.

We begin our investigation with a basic video generation model comprising a VQ-VAE~\cite{gray1984vq,oord2018vqvae} and an auto-regressive transformer. 
The raw task execution videos, collected from the environments described above, serve as the sole source of training data and hence the only source of knowledge. We convert video frames into discrete tokens using VQ-VAE and, similar to large language models (LLMs), train an auto-regressive transformer on these tokens under the next-token (or next-frame) prediction paradigm. During testing, the model generates new frames based on previous ones, and task-specific operations\textemdash such as moves in Go or robotic actions\textemdash are inferred from the newly generated frames (see Sec.~\ref{subsec:arch} for details). 

 Through this approach, we observe two key findings: 
 \begin{enumerate}
     \item  \textbf{The model can learn basic knowledge from raw videos}. This is evidenced by its ability to  learn Go rules and policies and fundamental robotic operations.  
     \item \textbf{The representation of visual change is crucial for knowledge learning}. While videos contain sufficient information for task completion, redundant representations of visual changes related to key decisions and actions hinder learning efficiency. A compact representation is essential for enhancing the model's learning efficiency and knowledge acquisitions. 
    
\end{enumerate}

\begin{figure}[t]
\includegraphics[width=\linewidth]{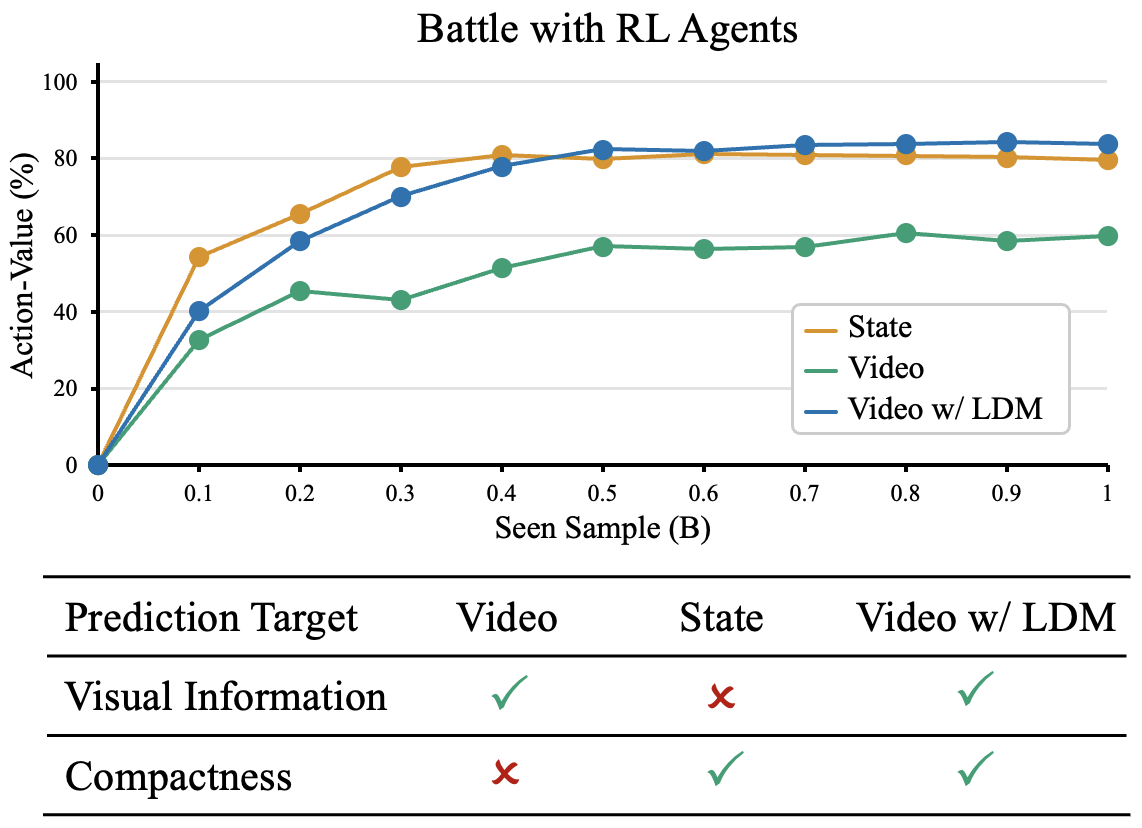}
\centering
\vspace{-2em}
\caption{\textbf{Comparison of prediction targets.} 
``State'', ``Video'' and ``Video w/ LDM'' refer to three different prediction targets: a state sequence (e.g., labeled positions of moves in Go), a raw video sequence, and a video sequence augmented with latent codes representing future visual changes (this approach is adopted by VideoWorld). “Action-Value” denotes the score for each move in the game, with details provided in Sec.~\ref{subsec:goeva}. By combining rich video information with a compact representation of visual changes, VideoWorld enables more effective learning.
}
\label{fig:fig1}
\vspace{-1.5em}
\end{figure}

Building on the observations above, we propose the  Latent Dynamics Model (LDM), which enhances both the efficiency and effectiveness of video learning while providing a mechanism to probe the model's learned knowledge. The LDM  compresses the future visual changes into a set of latent codes to serve as the compact representation of multi-step visual context.
% \textcolor{red}{[JS: explain the feature of 'latent code', like comprising multi-step information?]}. 
This allows the model to predict both video frames and latent code during training, improving its ability to capture and reason about diverse visual information, such as object interactions and scene dynamics.

These learned representations significantly enhance the model's learning efficiency and reasoning capability. As shown in Fig.~\ref{fig:fig1}, the video generation model with LDM achieves superior training efficiency and \emph{demonstrates a 5-dan\footnote{dan denotes the ranking system in Go, with higher levels indicating greater skill. 5-dan can already represent a highly trained human player.} professional level of performance against the RL agent KataGO\footnote{KataGO is an open-source implementation of AlphaGo and is arguably the strongest accessible Go agent.}}~\cite{wu2019katago}\textemdash an impressive accomplishment, given that this level is challenging even for most human players and our model achieves this with only 300 million parameters and solely through visual observation, without using search or reward learning techniques typical in RL~\cite{silver2016alphago, wu2019katago}. 

In the CALVIN and RLBench robotic scenarios, VideoWorld successfully performs various control tasks with promising data scaling behavior. It also demonstrates generalization capabilities across multiple environments, a challenge for RL methods due to environmental distinctions. Notably, it achieves performance close to the oracle models trained with ground-truth action labels. Above results highlight the model's ability to master complex tasks and acquire knowledge effectively.

Despite the promising results, we acknowledge that learning knowledge from unlabeled videos is still in its early stage. Improved visual representations and large-scale pretraining are likely to further enhance the model's learning capabilities.  In summary, we make following contributions:

\begin{itemize}
    \item We explore for the first time whether video generation models can learn sophisticated knowledge and observe two key findings: \textit{i}) merely observing videos suffices to learn complex tasks, and \textit{ii}) compact representations of visual changes greatly enhance knowledge learning.
    \item We propose VideoWorld, which leverages a latent dynamics model to represent multi-step future visual changes, thereby boosting both the efficiency and effectiveness of knowledge acquisition.
    \item We construct Video-GoBench, a large-scale video-based Go dataset for model training and evaluation, facilitating future research on knowledge learning from pure videos.
    
\end{itemize}

\section{Related Work}
\label{sec:related}

\subsection{Video Generation}
\label{subsec:videogen}
Existing video generation methods typically utilize Variational Autoencoders (VAEs)~\cite{kingma2013auto, van2017neural} to compress raw video data, and then perform large-scale generative pre-training in the compressed latent space based on diffusion~\cite{ho2020ddpm, song2021scorebased} or auto-regressive~\cite{yu2023magvit, pmlr-v119-chen20s} paradigms. Thanks to the scaling properties of the Transformer models~\cite{vaswani2017attention, peebles2023scalable}, the performance of video generation models can be continually improved, making significant progress in tasks such as text-to-video~\cite{videoworldsimulators2024, yang2024cogvideox,zhang2023controlvideo} and image-to-video~\cite{blattmann2023svd, guo2024i2v} generation. 

Recent works have begun to explore broader uses for video generation models. \cite{yang2024videodm} indicates that video generation models can be planners, agents, and environment simulators with the help of in-context learning~\cite{wang2023icl}, planning~\cite{huang2022planner}, and reinforcement learning~\cite{ouyang2022rlhf}. UniPi~\cite{du2023unipi} proposes that representing policies using text-conditioned video generation models enables effective learning of general-purpose decision-making systems. 

% Both studies underscore the importance of video generation in predicting real-world dynamics, positioning it not only as a tool for media entertainment but also as a critical component in developing general-purpose AI systems. 

However, existing works are limited to executing single, simple commands and heavily rely on language instructions. This work pioneers exploring AI models' ability to learn sophisticated reasoning knowledge solely from vision.

\subsection{Learning Knowledge}
\label{subsec:reasoning}
Learning knowledge in complex scenarios is crucial for intelligent agents, involving tasks like policy learning, planning, and decision-making, which are essential in applications such as game agents and autonomous driving. Existing works have achieved remarkable success in reasoning tasks using language models with techniques like chain-of-thought~\cite{wei2022cot, yao2024tree} and reinforcement learning~\cite{ouyang2022rlhf, lee2023rlaif}. Additionally, some studies focus on planning and reasoning within well-annotated, pre-defined environments. For example, ChessGPT~\cite{feng2024chessgpt} and Othello-GPT~\cite{li2023emergent} explore reasoning in chess using GPT-like models trained on annotated game data. Similarly, ChessBench~\cite{ruoss2024chessbench} introduces a dataset of 10 million annotated chess games to evaluate transformer models on planning tasks.

Existing methods typically rely on explicit state or action labels. In contrast, our work aims to explore knowledge learning in an unsupervised manner, driven purely by visual signals. Specifically, we investigate whether video generation alone can learn complex knowledge required for long-term, high-complexity tasks, without labeled data.

\subsection{Latent Actions}
\label{subsec:learn_with_la}
The concept most similar to our latent dynamics model is the latent action model. LAPO~\cite{schmidt2023lapo} learns latent actions unsupervised from videos, using an inverse dynamics model (IDM) to predict latent actions and a forward dynamics model (FDM) to ensure predictive consistency. Similarly, Genie~\cite{bruce2024genie} extracts latent actions through a causal latent action model and uses both video tokens and latent actions to predict the next frame. Additionally, LAPA~\cite{ye2024lapa} is an unsupervised pretraining method that teaches  Vision-Language-Action (VLA) models to learn discrete latent actions from videos without human-annotated labels. 
% LAPA overcomes the limitations of existing VLA models that rely on human annotations during pretraining, which often restricts the scale and diversity of data that can be used. 

In contrast, our work goes beyond action learning by showing that latent actions derived from unlabeled video data can support tasks requiring complex reasoning and planning. More importantly, we investigate how multi-step latent actions representing future visual changes enhance knowledge acquisition, especially in long-horizon tasks demanding extended planning and decision-making.

% \textcolor{blue}{this is only from the task angle. also include the difference of methods, we use latent representation for multiple future frames.}
% \textcolor{blue}{XJ: why? Directly list the difference in aim and method.}
% Furthermore, we delve deeper into how multistep latent actions can effectively enhance reasoning capabilities in long-scoped scenarios.

\begin{figure*}[t]
\includegraphics[width=\linewidth]{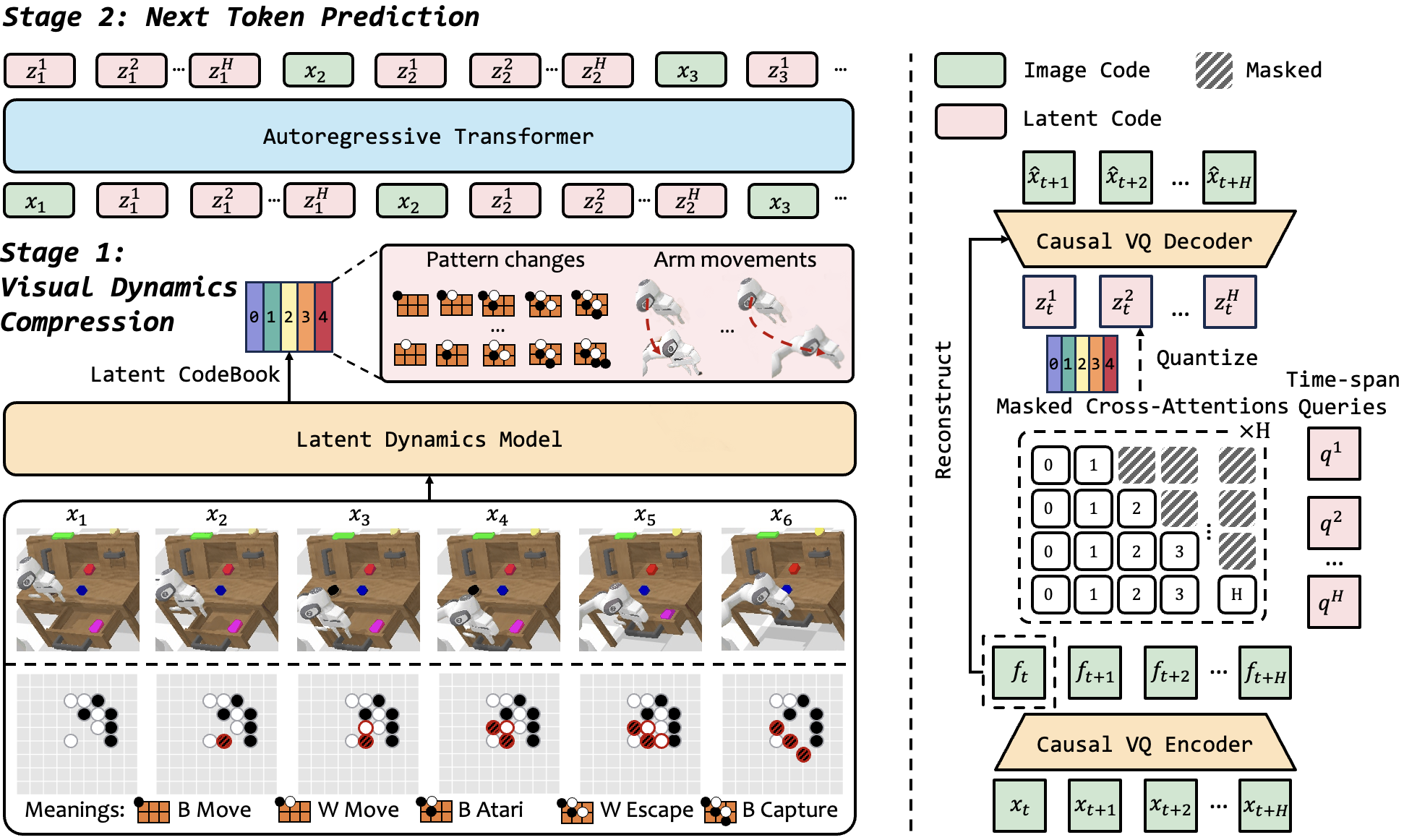}
\centering
\vspace{-2em}
\caption{\textbf{Overview of the proposed VideoWorld model architecture.} (Left) Overall architecture. (Right) The proposed latent dynamics model (LDM). First, LDM compresses the visual changes from each frame to its subsequent $H$ frames into a set of latent codes. Then, an auto-regressive transformer seamlessly integrates the output of LDM with the next token prediction paradigm. 
}
\label{fig:overview}
% \vspace{-1.5em}
\end{figure*}

\section{VideoWorld}
\label{sec:method}

\subsection{Basic Video Generation Framework}
\label{subsec:problem}
\noindent \textbf{Problem formulation.} We define the video generation process for learning knowledge in the context of a specific task as a tuple $\mathcal{G}=\left \langle \mathcal{X}, \mathcal{A}, \rho \right \rangle$, where $\mathcal{X}$ is the observation space, $\mathcal{A}$ is the action space, and $\rho$ is a video generator. An offline dataset $D=\{x_{1:T_n}^n\}_{n=1}^N$ consists of $N$ video demonstrations within this environment, with $T_n$ frames in each video sequence.
For brevity, we omit the subscript $n$ in the following text.  Our goal is to train a generator $\rho(x_{t+1} | x_{1:t})$ that models the conditional distribution of the next frame $x_{t+1}$, given the history of observations up to time step $t$ within observation space $\mathcal{X}$. 

Since the video frame sequences generated by $\rho$ inherently contain the necessary information for task progression, we can directly learn a task-relevant mapping function  $\pi$ that converts generated video frames into actions, Specifically,  $\pi(\cdot|x_{1:t+1}):\mathcal{X}\rightarrow\mathcal{A}$, where $\pi$ is a policy that maps the generated frames at time step $t+1$ to appropriate action in $\mathcal{A}$. This allows us to leverage the structure of generated videos to learn and execute the task without relying on explicit action annotations.

\noindent \textbf{Basic Framework.}  We focus on using an auto-regressive video generator to instantiate $\rho$. The basic framework includes a VQ-VAE encoder-decoder, and an auto-regressive transformer. The encoder converts video frames into discrete tokens, which the transformer uses for next-token prediction during training. During inference, the transformer generates discrete tokens for the next frame, which are subsequently converted back to pixel space by the decoder. 

In our implementation, the VQ-VAE is instantiated using a custom MAGVIT-v2~\cite{yu2024magvitv2}, equipped with an FSQ quantizer~\cite{mentzer2024fsq}, while the transformer is implemented using the Llama architecture~\cite{touvron2023llama}.  This setup allows us to generate high-quality frames, enabling effective modeling of visual sequences for diverse knowledge learning tasks.

\subsection{Learning with Latent Dynamic Model}
\label{subsec:arch}
Compared to a state sequence (e.g., move positions in Go), video provides essential visual information for understanding tasks\textemdash for instance, local patterns among stones in Go or environments in robotics. In preliminary experiments, we find that the basic framework can already learn fundamental knowledge solely from videos (see Sec.~\ref{subsec:findings}). However, as shown in Fig.~\ref{fig:fig1}, its learning efficiency still lags behind that of models trained on states. We attribute this to the inefficient representation of visual changes tied to critical decisions and actions. For example, while moves in Go can be encoded by just a few positional tokens in a state sequence, raw video requires significantly more tokens after passing through a vision encoder. This discrepancy adversely affects both learning efficiency and performance. Therefore, we develop VideoWorld, which integrates rich visual information with a compact representation of visual change for more effective video learning.

\begin{figure*}[t]
\includegraphics[width=\linewidth]{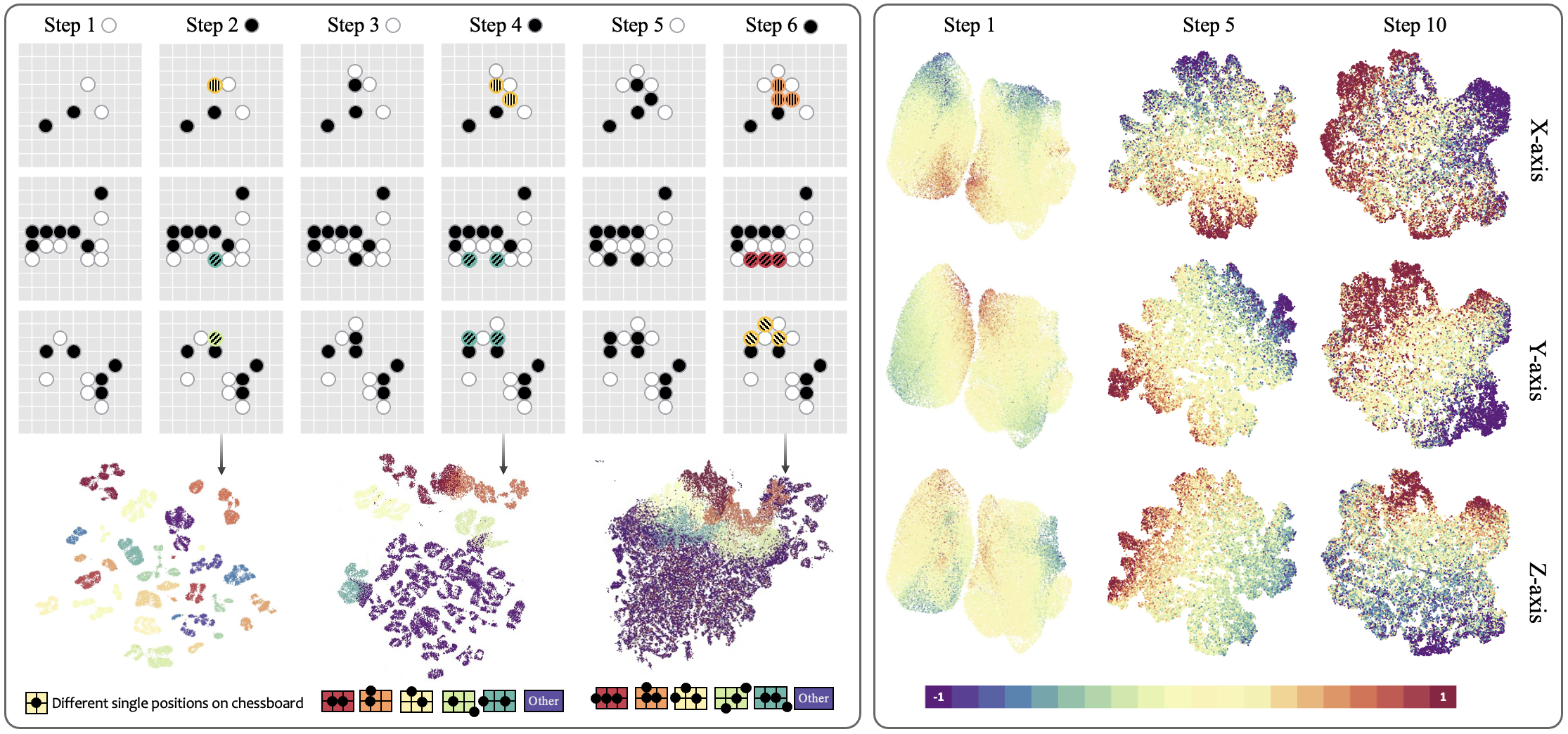}
\centering
\vspace{-2em}
\caption{\textbf{UMAP projection}~\cite{leland2018umap} \textbf{of the learned latent code} on the Go (Left) and CALVIN (right) training set. Each point represents
the continuous (pre-quantization) latent code generated by the LDM. In Go examples, odd steps represent white's moves, and even steps represent black's moves. We visualize the latent codes of black moves in steps 2/4/6. The legend shows examples of common patterns learned for new black moves. For clarity, these moves are highlighted on the board with added colors and lines to indicate new patterns. On the right, we visualize the latent codes of the robotic arm's movement along the X/Y/Z axes at intervals of 1, 5, and 10 frames. Points are color-coded by displacement range, with purple and red indicating the maximum displacement in opposite directions along each axis. } 
\label{fig:umap}
% \vspace{-1em}
\end{figure*}
\noindent \textbf{Latent Dynamics Model.} 
Video encoding typically requires hundreds or thousands of VQ tokens to capture the full range of visual information, leading to sparse embedding of knowledge across these tokens. To enhance efficiency, we introduce a latent dynamics model that uses query embeddings to represent visual changes across multiple frames.
%, specifically for reasoning and planning tasks. 
For example,   multi-step board changes in Go or continuous actions in robotics exhibit strong temporal correlations. By compressing these multi-step changes into compact embeddings, we not only increase the compactness of policy information but also encode guidance for forward planning.

The model employs a MAGVITv2-style~\cite{yu2024magvitv2} causal encoder-decoder, while intentionally omitting temporal downsampling to preserve detail in each frame. 
 For a video clip $x_{1:T}$, we sample each frame $x_t$ along with the subsequent $H$ frames, denoted as $x_{t+1:t+H}$. If fewer than $H$ steps remain, we apply replication padding.
In Fig.~\ref{fig:overview}, the encoder first extracts visual feature maps $f_{t:t+H}$ in a causal manner. Importantly, these features are not quantized, allowing them to retain detailed temporal information. 

Next, we define a set of attention blocks and corresponding learnable embeddings
$\{q^h\}_{h=1}^H$. Each query $q^h$, via the attention mechanism, captures change information in $f_{t:t+h}$, yielding a continuous latent representation   $\tilde{z}_t^h$, which is then quantized with a discrete codebook by FSQ~\cite{mentzer2024fsq}. This quantized representation serves as an information bottleneck, preventing the LDM from learning shortcuts (e.g. trivially copying $f_{t+h}$ to $z_t^h$).

Finally, the decoder uses the feature map $f_t$ and the latent change embeddings $\{z_t^h\}_{h=1}^H$ to predict the subsequent frames $\hat{x}_{t+1:t+H}$ in a causal manner. The training objective of LDM is minimize the $\ell_2$ distance between $x_{t+h}$ and $\hat{x}_{t+h}$. For further details on LDM, please refer to the Supp.

By sequentially encoding changes from $x_t$ to $x_{t+H}$ using multiple embeddings, we achieve a compact and informative representation of temporal dynamics, which is crucial for long-term reasoning and planning tasks. As shown in Fig.~\ref{fig:umap}, this model can learn meaningful embeddings that capture both short- and long-term dependencies in visual sequences.

\noindent \textbf{Auto-regressive Transformer.} The auto-regressive transformer is employed as the video generator,  and it seamlessly integrates the output of the LDM with the next token prediction paradigm. This integration offers several benefits, including improved modeling of long-term dependencies, enhanced reasoning capabilities, and more efficient learning of complex visual patterns over time. 

For each video $x_{1:T}$,  we first get latent codes $\{z_t^h\}_{t=1,h=1}^{T,H}$ using the latent dynamics model. These latent codes, along with discretized video frames are combined into a sequence for auto-regressive prediction. 
The codebook used by the video decoder is distinct from the one used in the latent dynamics model, and the vocabulary of the auto-regressive transformer is the union of both. This allows the transformer to leverage both the fine-grained visual details captured by the video encoder-decoder and the compact, task-relevant embeddings produced by LDM, enabling it to generate both visually coherent frames and maintain the underlying temporal dynamics captured by the LDM.

\noindent \textbf{\textbf{Mapping to Task Operation.}} During inference, at each time step $t$, we use the transformer to auto-regressively generate the latent codes  $\{\hat{z}_t^h\}_{h=1}^{H}$ and the predicted frame $\hat{x}_{t+1}$. To convert the generated content into actionable decisions for specific tasks, we further train an Inverse Dynamics Model (IDM) $\pi$ like~\cite{du2023unipi}.  IDM consists of several MLP layers and is trained independently from the video generator, using a small amount of video action label data. 

In the basic framework, the IDM takes the current frame   $x_t$  and the predicted frame  $\hat{x}_{t+1}$ to generate the corresponding action: $\pi(\cdot|x_t, \hat{x}_{t+1})$. When incorporating the LDM, the IDM is extended to take both the predicted frame and the latent codes, resulting in: $\pi(\cdot|x_t, \hat{x}_{t+1}, \{\hat{z}_t^h\}_{h=1}^{H})$. This allows the IDM to leverage the rich temporal representations encoded by LDM, enhancing its ability to produce more temporally consistent and accurate actions. For additional details, please refer to the Supp.~\ref{sec:supp_detail}.

\vspace{-0.5em}
\section{Video-GoBench}
\label{sec:gobench}
Our goal is to explore whether a deep generative model can learn complex knowledge from raw videos, including well-defined rules, advanced planning and reasoning. We prioritize tasks that emphasize high-level planning and reasoning while minimizing dependence on low-level visual details like shape and texture.

To this end, we introduce Video-GoBench, a purely visual Go benchmark, with a simple yet effective visual design to capture rule learning, reasoning, and planning, and enable direct evaluation through online gameplay. By simplifying visuals, we minimize distractions from intricate textures and dynamics, thereby allowing us to focus solely on assessing knowledge acquisition abilities. Additionally, the simulated Go environment allows for the seamless conversion of large-scale, state-based gameplay data into video sequences, supporting efficient and scalable evaluation.

\subsection{Dataset Generation}
To construct Video-GoBench, we collect 10 million 9x9 Go game records, including 3.2 million self-play training examples from  KataGo~\cite{wu2019katago} and 7.8 million from human-match games from OGS~\cite{ogs2024}. For the human data, we re-annotate the moves based on the original board states using KataGo to provide the optimal next moves for each state. The dataset comprises about 400 million unique board states, each converted into 256-pixel images. Additionally, we curated a test set of 1,000 matches, with each move re-annotated by KataGo. To improve its ability to benchmark model generalization on new boards, we removed opening moves\textemdash given the limited diversity in early-game positions\textemdash resulting in 56,000 unique board states.

\subsection{Evaluation}
\label{subsec:goeva}
We use the following metrics to evaluate our models:
% on various levels of knowledge learning and   measure training progress:

\begin{itemize}
    \item   \textbf{Legal rate}: The percentage of legal moves generated by the model on the test set, indicating the model’s ability to understand and adhere to the basic rules of Go.

   \item  \textbf{Game playing strength (Elo)}: Following \cite{ruoss2024chessbench}, we evaluate models through an internal tournament. Eight agents, as shown in Tab.~\ref{table:videogo}, play 400 games per pair, totaling 11.2K games. Elo ratings are calculated using BayesElo~\cite{coulom2008elo} with a default confidence of 0.5. Among the agents are three versions of the KataGo engine, fine-tuned to 1-dan, 5-dan, and 9-dan human skill levels, to showcase the model’s strategic levels in relation to human play. We anchor the KataGo 9-dan model’s Elo rating at 2700, aligning with the standard human 9-dan score.

   \item  \textbf{Action accuracy}: Percentage of test instances where the model selects the best move made by KataGO (oracle).

   \item \textbf{Action-Value}: During tournament matches, we record the board state at each move and use KataGo-9d to annotate the action-values, which are the expected cumulative rewards for all possible moves in each state, serving as oracle values. We then calculate the ratio of the action-value of each move made by the model to the oracle action-value. This metric measures the model's ability to evaluate moves. An action-value of 100\% indicates that the model's move matches the level of KataGo-9d.
\end{itemize}

% \vspace{-8em}
\section{Experiment}
\label{sec:experiment}

\begin{figure}[t]
\includegraphics[width=\linewidth]{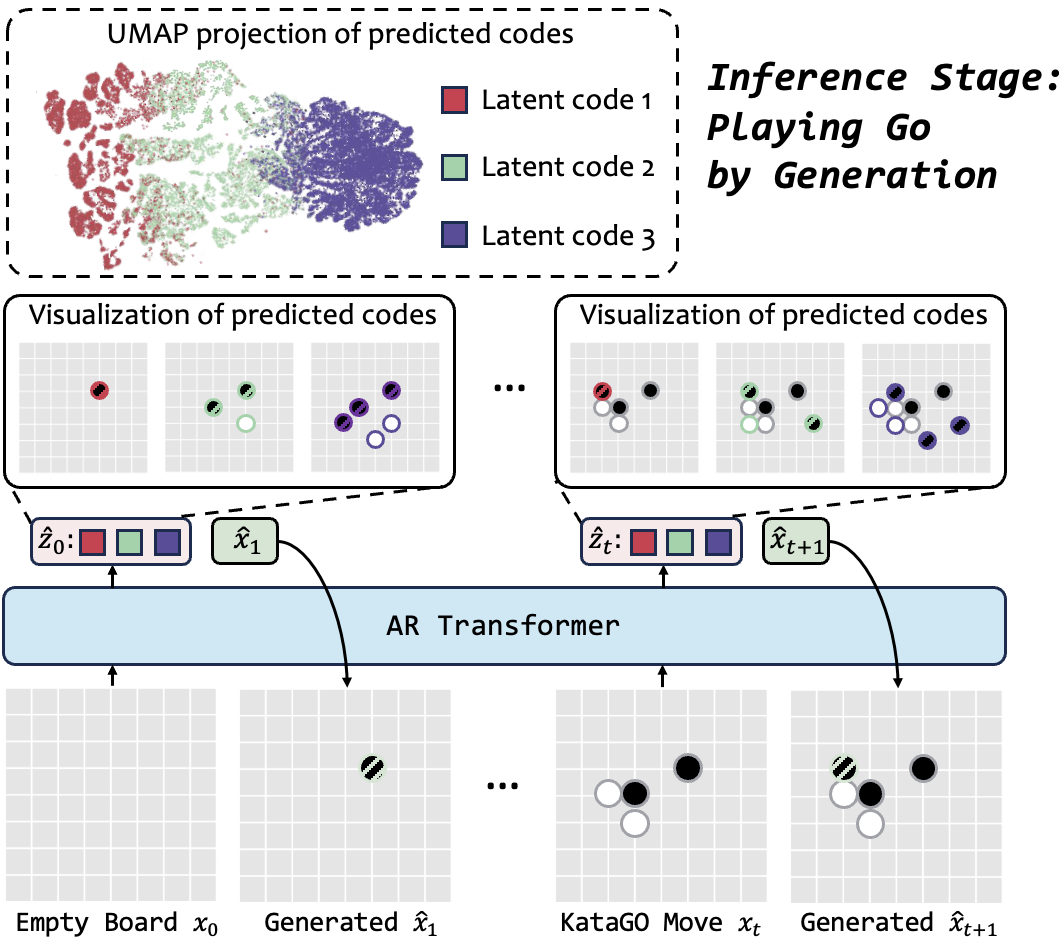}
\centering
\vspace{-2em}
\caption{\textbf{Illustration of playing against KataGO} and UMAP projection~\cite{leland2018umap} of the predicted latent code. Our model plays as black. The generated latent code is visualized through the LDM decoder and new stones in the visualization are marked with colors to match the legend. The visualization serves as a probe, indicating that the model shows signs of forward planning.   }
\label{fig:umap_test}
\vspace{-1.5em}
\end{figure}

\begin{table*}[t]
    \footnotesize
    \centering
    \begin{tabular}{p{0.1cm}lccccccc}
        \toprule
        Idx &Agent &Train &w/o Search &Input &Legal rate (\%) &Action-Value (\%) &Best Action Acc. (\%)  & Tournament Elo\\
        \midrule
        1&KataGO-human-1d  &RL &\XSolidBrush &State &100 &67.6 & 64.5 &2019\tiny{$\pm$23}  \\
        2&KataGO-human-5d  &RL &\XSolidBrush &State &100 &83.5 &83.7&2253\tiny{$\pm$20} \\
        3&{\color{lightgray}KataGO-human-9d (Oracle)}   &{\color{lightgray}RL} &{\color{lightgray}\XSolidBrush} &{\color{lightgray}State} &{\color{lightgray}100} &{\color{lightgray}100}&{\color{lightgray}100} &{\color{lightgray}2700} \\
        \midrule
        4&Transformer 300M  &SL &\Checkmark &State &99.8 &79.7&\underline{87.2}&\underline{2308}\tiny{$\pm$21} \\
        5&Transformer 300M  &SL & \Checkmark &Video&99.6&59.7&58.9&1998\tiny{$\pm$38} \\
        \midrule
        6&VideoWorld 50M (ours)    &SL & \Checkmark &Video &99.5 &73.9&80.9&2093\tiny{$\pm$25} \\
        7&VideoWorld 150M (ours) &SL & \Checkmark &Video &\textbf{99.7} &\underline{82.0}&86.7&2218\tiny{$\pm$23} \\
        8&VideoWorld 300M (ours) &SL & \Checkmark &Video &\textbf{99.7} &\textbf{83.7}&\textbf{88.1}&\textbf{2317\tiny{$\pm$25}} \\
        \bottomrule            
    \end{tabular}
\vspace{-1em}
\caption{\textbf{Comparison on Video-GoBench.} KataGO-human-9d represents the highest human level and serve as the Oracle for best actions.} 
% \vspace{-2em}
\label{table:videogo}   
\end{table*}
% \begin{center}
% \begin{table}[t]
%     \footnotesize
%        \centering
%         \resizebox{\columnwidth}{!}
%         {
%         \begin{tabular}{lcccc}
%             \toprule
%              \multirow{2}*{Agents}   &\multirow{2}*{Input/Ouput} & \multicolumn{3}{c}{Task Success Rate (\%)}\\
             
%              \specialrule{0em}{0pt}{1pt}
%              \cline{3-5}
%              \specialrule{0em}{0pt}{1pt}
%             &&Push &Open/Close &Turn on/off\\

%             \specialrule{0em}{0pt}{1pt}
%             \hline
%             \specialrule{0em}{0pt}{1pt}
            
%             {\color{lightgray}Transformer (Oracle)} &{\color{lightgray}Video\&GT Action} &{\color{lightgray}75.4} &{\color{lightgray}95.3} &{\color{lightgray}96.2} \\
%             Transformer  &Video &17.3 &24.1 &19.2 \\

%              \specialrule{0em}{0pt}{1pt}
%             \hline
%             \specialrule{0em}{0pt}{1pt}
       
%              VideoWorld &Video &56.2 &75.4 &72.1 \\
%             \bottomrule            
%         \end{tabular}
%         }
%     \caption{\textbf{Comparison on CALVIN benchmark.}}  
%     \label{table:calvin}
% \end{table}
% \end{center}

\begin{table}[t]
    \footnotesize
    \centering
    \resizebox{\columnwidth}{!}{
    \begin{tabular}{lcccc}
        \toprule
        \multirow{2}*{Agents}   &\multirow{2}*{Input/Ouput} & \multicolumn{3}{c}{Task Success Rate (\%)}\\
        \cmidrule(lr){3-5}
        &&Push &Open/Close &Turn on/off\\
        \midrule  
        MCIL~\cite{mees2022matterslanguageconditionedrobotic}  &Video/Lab. Action &33.0 &38.7 &41.2 \\
        HULC~\cite{lynch2021languageconditionedimitationlearning}  &Video/Lab. Action &65.8 &80.9 &85.3 \\
        {\color{lightgray}Transformer (Oracle)} &{\color{lightgray}Video/Lab. Action} &{\color{lightgray}75.4} &{\color{lightgray}95.3} &{\color{lightgray}96.2} \\
        Transformer  &Video &17.3 &24.1 &19.2 \\
        \midrule
        VideoWorld &Video &56.2 &75.4 &72.1 \\
        VideoWorld (+10k data)$^{\dagger}$ &Video &65.3 &81.2  &79.3\\
        VideoWorld (+30k data)$^{\ddagger}$ &Video &72.7 &91.0 &93.8   \\
        \bottomrule            
    \end{tabular}
    }
\vspace{-1em}
\caption{\textbf{Comparison on CALVIN benchmark.} ``Lab.'' means annotated labels. All models have 300M parameters. $^{\dagger}$ and $^{\ddagger}$ denote using an additional 10k and 30k CALVIN trajectories for training, respectively.}  
% \vspace{-1em}
\label{table:calvin}
\end{table}

\begin{table}[t]
    \footnotesize
    \centering
    \resizebox{\columnwidth}{!}{
    \begin{tabular}{lccccc}
        \toprule
        \multirow{2}*{Agents}  & \multicolumn{3}{c}{CALVIN} & \multicolumn{2}{c}{RLBench}\\
        \cmidrule(lr){2-4} \cmidrule(lr){5-6}
        &Push &Open/Close &Turn on/off &Microwave &Fridge\\
        \midrule  
        {\color{lightgray}Transformer (Oracle)} &{\color{lightgray}61.3} &{\color{lightgray}79.5} &{\color{lightgray}78.0} &{\color{lightgray}72.1} &{\color{lightgray}69.0} \\
        Transformer  &6.5 &13.0 &15.6 &12.0 &10.9\\
        % \midrule
        {VideoWorld} &56.0 &74.8 &74.5 &67.1 &62.5   \\
        \bottomrule            
    \end{tabular}
    }
\vspace{-1em}
\caption{\textbf{Results of joint training on CALVIN and RLBench.}  }  
% \vspace{-2.5em}
\label{table:calvinrlbench}
\end{table}
\begin{table*}[t]
\centering
%#################################################
% MAE decoder depth
%#################################################
\subfloat[
Different Compression length on Go.
\label{tab:abla_timespan}
]{
\centering
\begin{minipage}{0.29\linewidth}{\begin{center}
\tablestyle{5pt}{1}
\begin{tabular}{c cc}
    % \toprule
    \multicolumn{1}{c|}{\multirow{2}*{\makecell{Compression \\ length $H$}}}  &\multicolumn{2}{c}{Go}    \\
      \multicolumn{1}{c|}{} &Act-Value & Act-Acc.   \\
    \shline
    \specialrule{0em}{0pt}{1pt}
    \multicolumn{1}{c|}{baseline} &47.5  &44.3 \\
    % baseline &31.8  &40.6  &55.1 &56.0\\
    % + Higher resolution &40.1  &47.2 &64.3 &65.6\\
    
    \multicolumn{1}{c|}{1} &70.3  &77.0  \\
    \multicolumn{1}{c|}{3} &\underline{72.5}  &\underline{80.6} \\
    % \hline
    \multicolumn{1}{c|}{5} &\textbf{73.9}  &\textbf{80.9} \\
    && \\
\end{tabular}
\end{center}}
\end{minipage}
}
\hspace{1em}
%#################################################
% MAE decoder width
%#################################################
\subfloat[
Different Compression length on CALVIN.
\label{tab:abla_timespan_calvin}
]{
\begin{minipage}{0.33\linewidth}{\begin{center}
\tablestyle{5pt}{1}
\begin{tabular}{c  ccc}
    % \toprule
    \multicolumn{1}{c|}{\multirow{2}*{\makecell{Compression \\ length $H$}}}  &\multicolumn{3}{c}{CALVIN}    \\
      \multicolumn{1}{c|}{} & Push & Open/Close &Turn On/Off   \\
    \shline
    \specialrule{0em}{0pt}{1pt}
    \multicolumn{1}{c|}{baseline} &12.7  &20.8 &15.6 \\
    % baseline &31.8  &40.6  &55.1 &56.0\\
    % + Higher resolution &40.1  &47.2 &64.3 &65.6\\
    
    \multicolumn{1}{c|}{1} &33.7  &53.6 &67.3  \\
    \multicolumn{1}{c|}{5} &\underline{46.8}  &\underline{66.1} &\underline{69.6} \\
    % \hline
    \multicolumn{1}{c|}{10} &\textbf{50.3}  &\textbf{71.1} &\textbf{69.7} \\
    &&& \\
\end{tabular}
\end{center}}
\end{minipage}
}
\hspace{1em}
%#################################################
% MAE with mask token on encoder
%#################################################
\subfloat[
Intervene latent codes with different index
\label{tab:abla_intervene}
]{
\begin{minipage}{0.29\linewidth}{\begin{center}
\tablestyle{5pt}{1}
\begin{tabular}{c | cc }
    % \toprule
     \multirow{2}*{\makecell{Code \\ Index}} &\multicolumn{2}{c}{Go}  \\
     &Act-Value  &Act-Acc.    \\
    \shline
    \specialrule{0em}{0pt}{1pt}
    None  &\textbf{73.9}  &\textbf{80.9}      \\
    1 &46.2  &42.1    \\

    2 &69.5  &76.6     \\
    
    3  &\underline{72.1}  &\underline{80.6}     \\
    All  &45.8  &43.7   \\
\end{tabular}
\end{center}}
\end{minipage}
}
\\
\centering
% \vspace{-0.5em}
%#################################################
% MAE targets
%#################################################
\subfloat[
Different codebook size of LDM.
\label{tab:abla_codebooksize}
]{
\begin{minipage}{0.49\linewidth}{\begin{center}
\tablestyle{5pt}{1}
 \begin{tabular}{ c | cc | ccc}
                % \toprule
                  \multirow{2}*{\makecell{LDM \\ codebook  size  }}  \multirow{2}*  & \multicolumn{2}{c|}{Go} &\multicolumn{3}{c}{CALVIN} \\
                 
                     &Act-Value  &Act-Acc. &Push  &Open/Close &Turn On/Off  \\
                \shline
                \specialrule{0em}{0pt}{1pt}
                  729 &65.5   &71.1 &12.9 &20.0 &16.0     \\
                   15,625&\underline{69.3}   &\underline{72.8} &20.5 &24.1 &\underline{36.3}    \\
                  64,000&\textbf{73.9}   &\textbf{80.9} &\textbf{50.3}  &\textbf{71.1} &\textbf{69.7}   \\
                  262,144&50.1  &53.2  &\underline{29.8} &\underline{30.0} &31.7 \\
                % 4 & \Checkmark & \Checkmark & \Checkmark & 52.9 & \textbf{54.0} \\
                % \bottomrule
            \end{tabular}
\end{center}}
\end{minipage}
}
\hspace{1em}
% %#################################################
% % MAE data aug
% %#################################################
% \subfloat[
% \textbf{Data augmentation}. Our MAE works with minimal or no augmentation.
% \label{tab:aug}
% ]{
% \centering
% \begin{minipage}{0.29\linewidth}{\begin{center}
% \tablestyle{4pt}{1.05}
% \begin{tabular}{y{54}x{22}x{22}}
% case & ft & lin \\
% \shline
% none & 84.0 & 65.7 \\
% crop, fixed size & 84.7 & 73.1 \\
% crop, rand size & \baseline{\textbf{84.9}} & \baseline{\textbf{73.5}} \\
% crop + color jit & 84.3 & 71.9 \\
% \end{tabular}
% \end{center}}
% \end{minipage}
% }
% \hspace{1em}
%#################################################
% MAE with mask types
%#################################################
\subfloat[
Data source ablation. $^\dagger$ means human data without re-annotation.
\label{tab:abla_data}
]{
\begin{minipage}{0.45\linewidth}{\begin{center}
\tablestyle{5pt}{1}
\begin{tabular}{l | c|cc }
    % \toprule
    \multirow{2}*{\makecell{Data \\ source}} &\multirow{2}*{\makecell{Data \\amount}} &\multicolumn{2}{c}{Go}  \\
    &  & Act-Value & Act-Acc.  \\
    \shline
    \specialrule{0em}{0pt}{1pt}
    Human$^\dagger$ &6.8M &34.6 &34.8  \\
    Human &6.8M &\underline{71.5} &75.7  \\
    % \specialrule{0em}{0pt}{1pt}
            % \hline
            % \specialrule{0em}{0pt}{1pt}
    KataGo &3.2M &70.8 &\underline{77.5} \\

     All &10M &\textbf{73.9} &\textbf{80.9}  \\
\end{tabular}
\end{center}}
\end{minipage}
}
%#################################################
% \vspace{-0.5em}
\caption{\textbf{Ablations and analysis.} We conduct all experiments based on our 50M model.}
\vspace{-1em}
\label{tab:ablations} %\vspace{-1.5em}
\end{table*}
% \begin{table}[t]
%     \footnotesize
%     \centering
%     % \tabcolsep=0.18cm
%     {
    
% \begin{tabular}{c| cc}
%     % \toprule
%     \multirow{2}*{\makecell{Prediction \\ Target}}    &\multicolumn{2}{c}{Go}   \\
%       &Act-Value & Act-Acc.   \\
%     \shline
%     \specialrule{0em}{0pt}{1pt}
%     video &47.5  &44.3 \\
%     % baseline &31.8  &40.6  &55.1 &56.0\\
%     % + Higher resolution &40.1  &47.2 &64.3 &65.6\\
    
%     code &70.3  &77.0  \\
%     code/video &\underline{72.5}  &\underline{80.6} \\
%     % \hline

% \end{tabular}
% }
% \caption{\textbf{Latent code prediction only.}}
% \label{table:supp_onlylc_go}
% % \vspace{-1em}
% \end{table}

\begin{table}[t]
    \footnotesize
    \centering
    \resizebox{\columnwidth}{!}{
    
\begin{tabular}{c|cc|ccc}
    % \toprule
    \multirow{2}*{\makecell{Prediction \\ Target}}    &\multicolumn{2}{c|}{Go} &\multicolumn{3}{c}{CALVIN}   \\
      &Act-Value & Act-Acc. &Push & Open/Close &Turn on/off   \\
    \shline
    \specialrule{0em}{0pt}{1pt}
    video &47.5 &44.3 &12.7  &20.8 &15.6 \\
    % baseline &31.8  &40.6  &55.1 &56.0\\
    % + Higher resolution &40.1  &47.2 &64.3 &65.6\\
    
    code &73.0  &78.6 &47.2 &70.0 &65.1 \\
    code/video &73.9  &80.9 &50.3 &71.1 &69.7 \\
    % \hline
    % \bottomrule           
\end{tabular}
}
\caption{\textbf{Latent code prediction only} with 50M parameters.}
\label{table:supp_onlylc_go}
% \vspace{-2em}
\end{table}

\subsection{Implementation Details}
Our auto-regressive transformer is randomly initialized, with its encoder initially trained on the target dataset for reconstruction and then frozen. Trainable parameters including the transformer layers, projection layers, and output layers. The default vocabulary size for both LDM and transformer encoders is 64,000, corresponding to FSQ levels $[8,8,8,5,5,5]$.
For Go and CALVIN environments, we increase the number of downsampling layers in each encoder, compressing each frame to 4x4. We use a frame length of 6 for Go and 10 for CALVIN. We train and test in the CALVIN ABCD$\rightarrow$D split. RLBench shares the same training settings as CALVIN, with 20k trajectorie data generated by custom scripts. The training uses the AdamW~\cite{loshchilov2019adamw} optimizer with a learning rate of 0.0003 and no weight decay. The batch size is set to 256 for Go and 32 for CALVIN, requiring approximately 4 and 2 days of training on 8 A100 GPUs, respectively.

\subsection{Benchmarks}
We evaluate VideoWorld on three benchmarks: Video-GoBench, CALVIN and RLBench. Go tests long-horizon reasoning and complex forward planning, while the latter two involve more intricate visual information.
For Go, we use the evaluation metric detailed in Sec.~\ref{subsec:goeva}. In CALVIN benchmark, our model controls a Franka Emika Panda robot with a parallel-jaw gripper in an environment with a desk, openable drawer, colored blocks, and an LED and light bulb. We evaluate three tasks: Push Blocks, Open/Close Drawer, and Turn On/Off Light. Each task is specified by an instruction label, such as ``{\tt go push the red block right}'',  which is provided to the transformer as a condition for video generation. In RLBench benchmark, the model controls the same robotic arm, but the environment and camera views differ from CALVIN. We evaluate two tasks: Close Microwave and Close Fridge. Object positions in both environments are randomly generated for each execution. We sample 500 tasks randomly and record the success rate.

\subsection{Key Findings with The Basic Framework} 
\label{subsec:findings}

% \vspace{-15mm}
\noindent \textbf{Model can learn basic knowledge from raw videos.} As shown in Tab.~\ref{table:videogo} (idx 5) and Tab.~\ref{table:calvin} (line two), using the basic framework to train on Go videos and robotic videos enables the model to master the rules of Go, and learn basic robotic operation skills. On the Go test set, the model demonstrates almost 100\% legality, indicating that it has mastered rules including prohibitions against repeated moves, suicide moves, and more advanced rules like ko fights. Additionally, it achieves nearly a 50\% accuracy rate in predicting the best moves, showing a moderate understanding of Go strategy. In robotic scenarios, the model also demonstrates a certain ability to complete tasks. Compared to Go, handling more complex appearance without relying on pre-training imposes higher demands on the model's generation quality.

\noindent \textbf{Representation of visual change is crucial.} While the basic framework can learn from video, we find that using more compact representations of visual changes yields significantly better performance and efficiency. In Go, a state sequence encodes each move and its corresponding positional token to represent changes on the board; in contrast, raw video captures additional visual details beyond these key actions, thus requiring a substantially larger number of tokens. Although the state-based approach is deficient in representing local patterns and shapes among stones\textemdash an important property in Go\textemdash its compactness leads to higher learning efficiency and better results. As shown in Tab.~\ref{table:videogo}, training with these state sequences significantly improves performance across all metrics (idx 4 vs. 5) under the same data budget. Fig.~\ref{fig:fig1} further demonstrates that the state representation converges faster. In robotic scenarios, although fully representing the environment with state sequences may not be feasible, adding action labels for each frame still markedly boosts task completion rates. These findings underscore the importance of a compact representation of visual change for effective video learning.

\subsection{Results with LDM} 
\label{subsec:res_ldm}
Based on the above observations, we introduce the Latent Dynamics Model (LDM) to enhance the efficiency and effectiveness of video learning, resulting VideoWorld model.

\noindent \textbf{Results on Go.}
In Tab.~\ref{table:videogo}, we compare VideoWorld with three official KataGo models calibrated to human skill levels: 1-dan, 5-dan, and 9-dan. The 1-dan model represents players with a foundational understanding of Go, while the 5-dan and 9-dan levels indicate progressively advanced skills, with 9-dan serving as an Oracle at the highest human level. Notably, all these models are RL-based, using search and reward mechanisms.

We evaluate VideoWorld at three parameter scales: 50M, 150M, and 300M. All models show strong generalization on novel boards. Even the smallest model, VideoWorld-50M surpasses KataGo-1d (Elo 2093 vs. 2019), while VideoWorld-300M outperforms the human-level KataGo-5d (Elo 2317 vs. 2253), demonstrating VideoWorld’s strong capability for mastering complex tasks. Compared to its counterpart without LDM, VideoWorld shows a significant improvement (Elo 2317 vs. 1998), underscoring the benefits of learning with latent representations.

Furthermore, we observe that performance across all metrics improves consistently with model size, suggesting that VideoWorld’s knowledge learning capabilities continue to scale effectively, with potential for further enhancements as model size increases.

\noindent {\textbf{Results on CALVIN.}} Tab.~\ref{table:calvin} presents the results of the models on the CALVIN benchmark. VideoWorld, relying solely on observed videos and does not use action labels, achieves performance close to that of models supervised with real action labels. This demonstrates that our proposed LDM effectively supports video-based knowledge learning, even in more visually complex scenarios. Furthermore, we use a supervised learning agent~\cite{wu2023gr1} to generate an additional 30k trajectory data in the CALVIN environment, which we incorporate into VideoWorld's training. As shown in Tab.~\ref{table:calvin}, VideoWorld demonstrates data scaling capabilities, further approaching oracle performance as data volume increases. This demonstrates the potential of our method for application in larger-scale training.

\noindent {\textbf{Generalization across multiple environments.}} To further validate VideoWorld's generalization across different environments, we introduce two additional tasks from RLBench, which are visually distinct from CALVIN: closing microwave and closing fridge. For each task, we generate 10k trajectories and train VideoWorld on these data combined with CALVIN. We then evaluate the task success rate in both environments. As shown in Tab.~\ref{table:calvinrlbench}, VideoWorld generalizes well across two settings, simultaneously mastering skills in two environments and approaching oracle performance. Reinforcement learning methods often struggle to generalize across different environments, as they heavily rely on task-specific states/action/rewards, etc. This also demonstrates the potential of our method to serve as a general knowledge learner. See Supp.~\ref{sec:supp_detail} for more details.

\subsection{Understanding Learned Knowledge with LDM}
\label{subsec:understand_ldm}
The latent representation learned in LDM provides valuable insights into the knowledge learning process of VideoWorld. Below we offer an in-depth analysis of what our model learns through latent representations.
% \begin{enumerate} 

\noindent {\textbf{LDM learns patterns in the training set.}} As shown in Fig.~\ref{fig:umap}, the latent codes on the training set capture both short- and long-term dependencies, demonstrating the model's ability to represent knowledge at different temporal scales. In the Go scenario, salient regions in the latent codes correspond to common move patterns, indicating that the model effectively embeds multi-step strategies into a compressed space, hence aiding decision-making and reasoning. Similarly, in the robotics scenario, the clustering of latent codes across steps reveals key dynamic dependencies over various time ranges, thus benefiting diverse manipulation tasks.

\noindent {\textbf{LDM enables forward planning during testing.}} We examine the role of codes during inference. The visualization in Fig.~\ref{fig:umap_test} shows that codes from different steps group by output positions, suggesting that VideoWorld models long-range changes progressively, similar to human forward-planning. The visualization also includes imagination of the opponent's moves, achieving a high average action-value of 71.2\% and action accuracy of 74.3\%. This indicates that, at each step, VideoWorld considers long-term changes in the game situation within the latent space, enabling it to make strategic moves with a long-term perspective. 

\begin{figure}[t]
\includegraphics[width=\linewidth]{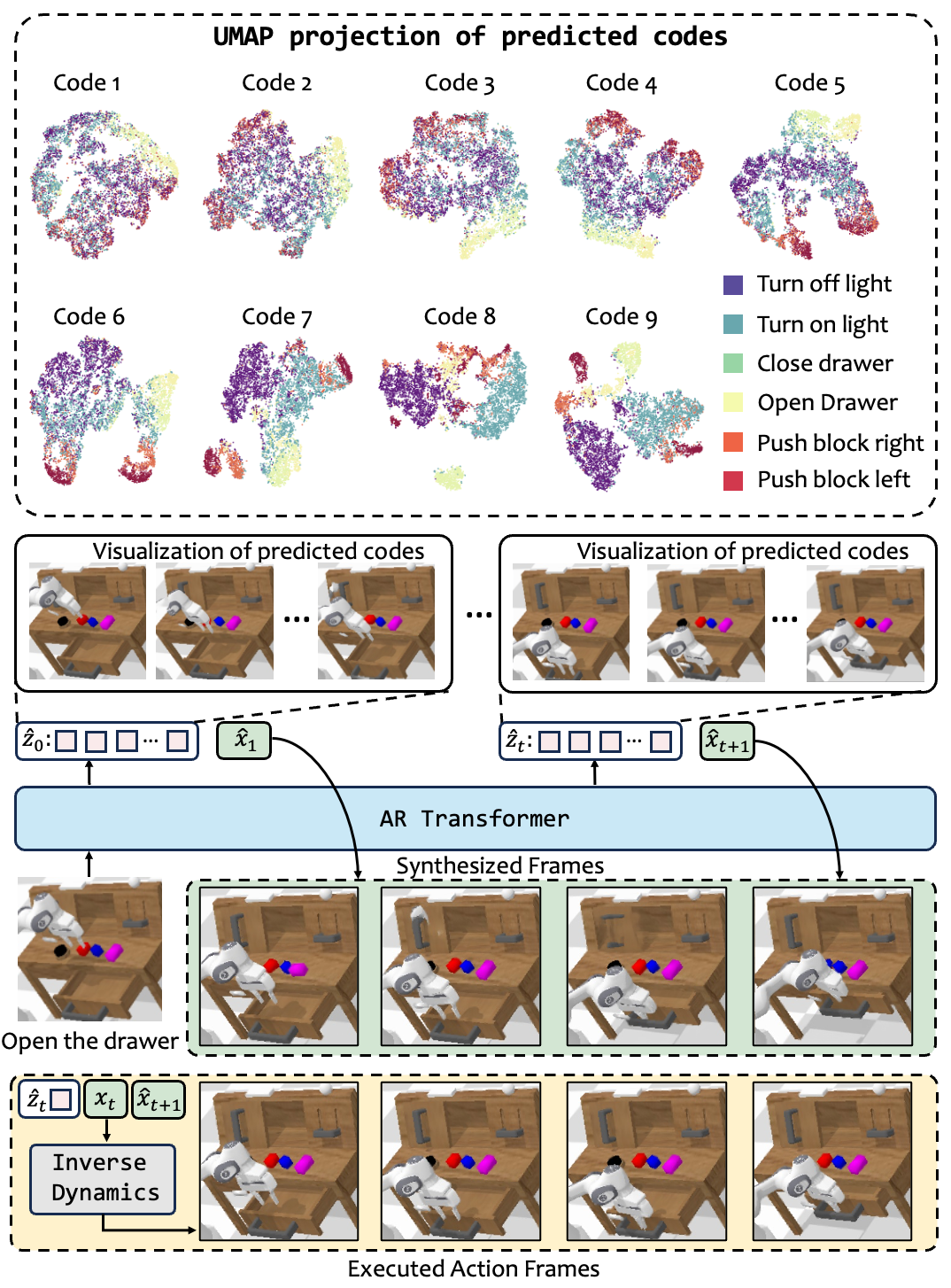}
\centering
\vspace{-1em}
\caption{\textbf{Illustration of robotic manipulation} and UMAP projection of the predicted latent code during inference. Latent codes are visualized through the LDM decoder. The UMAP projection illustrates the 9 predicted latent codes (i.e. $H=9$) across different tasks, with each point color-coded by task type. Visualizations with a yellow background show the model's actual robotic arm control during inference, while those with a green background represent the model's next-frame predictions during training.}
\label{fig:umap_test_calvin}
% \vspace{-0.5em}
\end{figure}

Similar findings are observed in the robotic scenario. We visualize the predicted latent codes during inference across different tasks in Fig.~\ref{fig:umap_test_calvin}. Here, $H=9$, meaning the transformer generates 9 latent codes per time step, corresponding to 9 prediction steps. As shown, the latent codes for different prediction steps are grouped by task type, indicating that they capture task-relevant dynamics. Codes for steps 1–4 show greater overlap, likely because they focus on fine-grained displacements shared across tasks. In contrast, steps 5–9 show more distinct separation by task type, highlighting the model's ability to progressively capture long-range changes specific to each task.
% Supp.~\ref{subsec:code_ana} shows a similar pattern in robotic scenario.

\noindent {\textbf{LDM generates causally interrelated codes.}} To further investigate the impact of the learned latent codes, we conduct an intervention experiment in Tab.~\ref{tab:abla_intervene}. We replace latent codes at different time steps with random tokens and observe the effect on performance. Intervening with the first code has the greatest effect, likely due to the causal dependencies learned among the codes. This aligns with intuition: altering the first code, which represents the immediate next optimal decision, affects all future decisions over a longer horizon. These results highlight the importance of the order and structure of latent codes for effective reasoning.

\subsection{More Ablations}

\noindent \textbf{Horizon length in latent dynamics model.} Tab.~\ref{tab:abla_timespan} shows the effects of varying compression lengths during LDM training. 
We use a default compression length of 5 steps for Go and 10 steps for CALVIN.
% For Go, we use a default compression length of 5 steps, and for CALVIN, we use 10 steps.
When we vary the compression length while keeping the codebook size constant, we observe that with a compression length of 1 step (similar to pseudo action labels), both Go and CALVIN improvement significantly over the baseline. For Go, optimal performance is at 5 steps, but further increasing the length causes the LDM training to fail to converge. We attribute this to the exponential growth in possible board position variations, which overwhelms the codebook and causes training instability. In contrast, the simpler dynamic changes in CALVIN environment allow for longer compression lengths.

% \noindent \textbf{Intervene latent visual process codes.} 

\noindent \textbf{Latent codebook size.}    Tab.~\ref{tab:abla_codebooksize} examines the impact of LDM's codebook size on knowledge learning. By adjusting the FSQ level, we test four different codebook sizes to evaluate how this affects the model's ability to learn effective policies. In the Go scenario, we find that even a smaller vocabulary yields significant improvements; however, this is not the case for the robotic scenario. This discrepancy likely arises because the robotic scenario involves more motion and appearance information, requiring a larger vocabulary to capture essential details. When the vocabulary size becomes too large, LDM training struggles to converge, leading to a noticeable performance drop in both scenarios.

\begin{figure}[t]\centering
  \subfloat[\label{fig:supp_vis_cap}Capture opponent's stones]{\includegraphics[width=\linewidth]{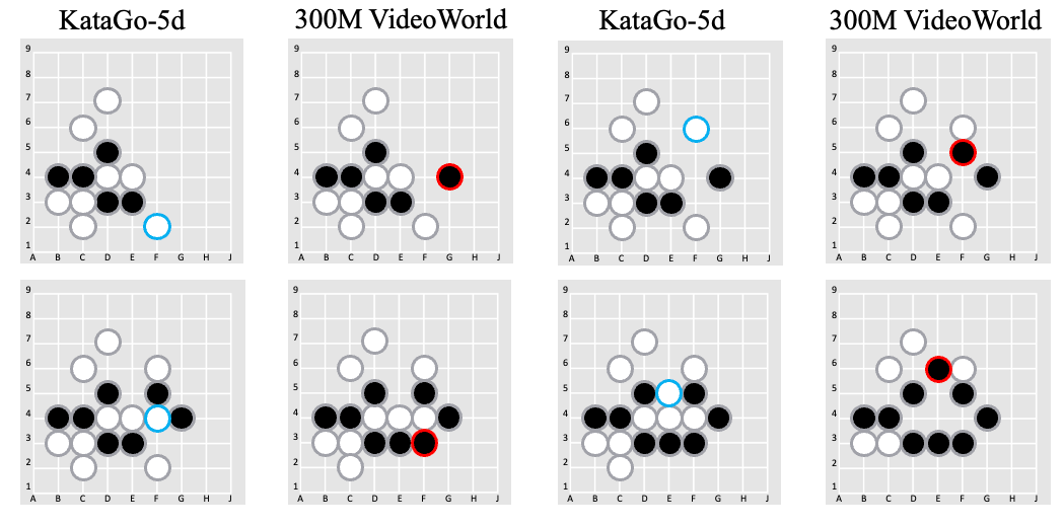}  }\\
% \hfil
% \vspace{-0.5em}
\subfloat[\label{fig:supp_vis_sac}Sacrificing short-term gains to capture more opponent stones.]{\includegraphics[width=\linewidth]{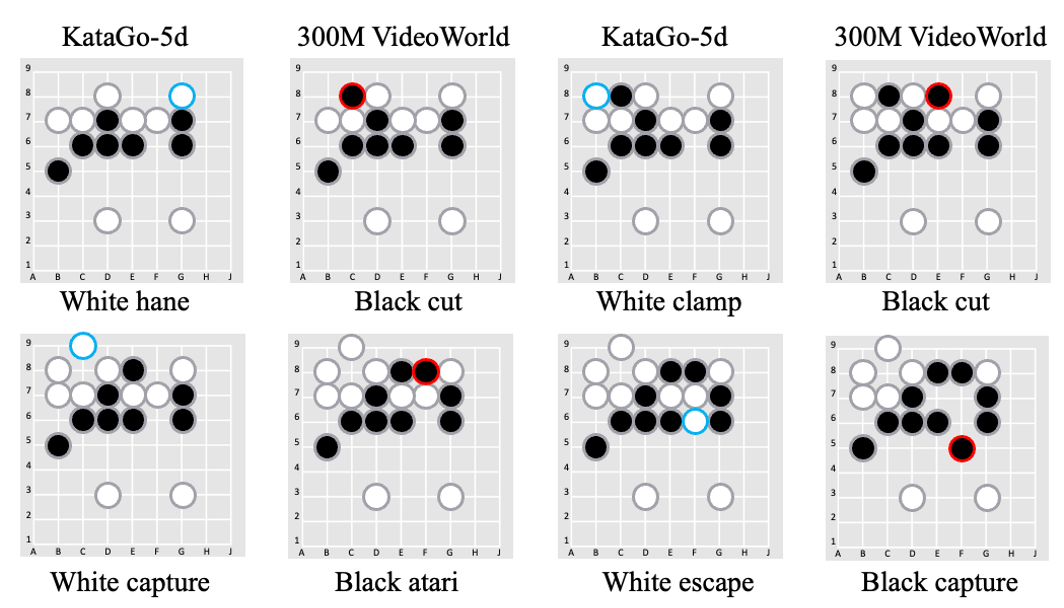} }
\hfill
\vspace{-1em}
\caption{\textbf{Visualizations of learned Go Strategies}. our model captures opponent stones using the squeeze tactic and self-sacrifice tactic. New black stones are red, new white stones are blue.}
\vspace{-1em}
\label{fig:supp_vis_go}
\end{figure}

\begin{figure}[t]
\includegraphics[width=\linewidth]{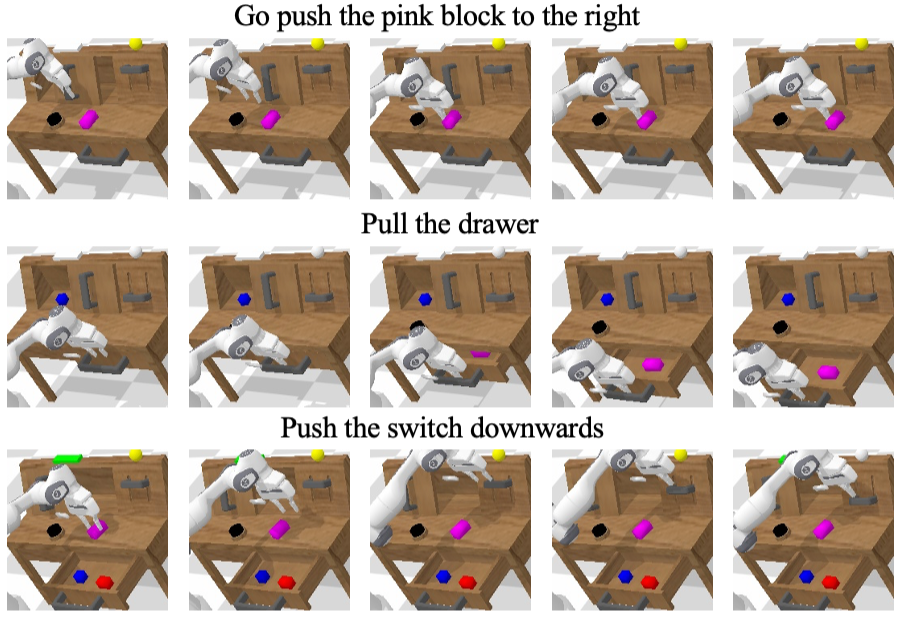}
\centering
\vspace{-2em}
\caption{\textbf{Visualizations of performing CALVIN tasks.}}
\label{fig:supp_vis_calvin}
% \vspace{-1em}
\end{figure}
% Visualizations of VideoWorld performing robotic manipulation task} in CALVIN environment.

\begin{figure}[t]
\includegraphics[width=\linewidth]{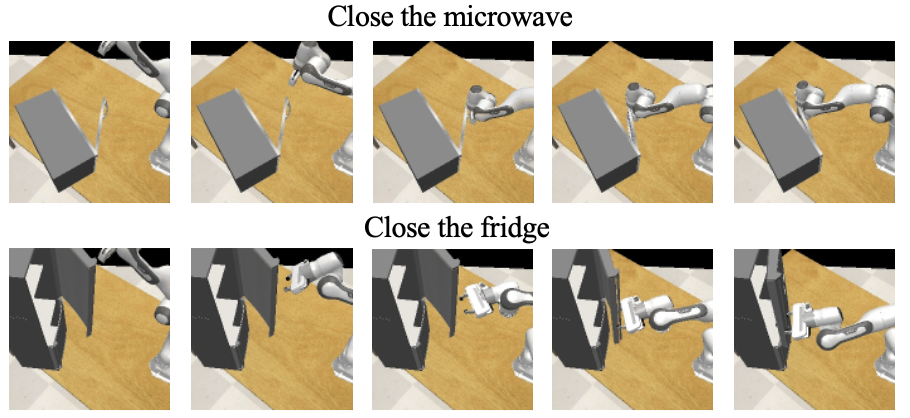}
\centering
\vspace{-2em}
\caption{\textbf{Visualizations of performing RLBench tasks.}}
\label{fig:supp_vis_rlbench}
\vspace{-1em}
\end{figure}

\noindent \textbf{Data quality.} Tab.~\ref{tab:abla_data} examines the effect of data quality on learning. Video-GoBench dataset has two types of data: human matches and KataGO self-play matches. These data sources have different distributions, and their policy content varies, potentially influencing the learning process. We find that using the original human data significantly impacts the final performance, as it includes players of varying skill levels, highlighting the importance of data quality.

\noindent \textbf{Latent code prediction only.}
To demonstrate the necessity of  jointly predicting the latent code $\{\hat{z}_t^h\}_{h=1}^H$ and next frame $\hat{x}_{t+1}$ given $x_{1:t}$, we remove the supervision for the next frame $\hat{x}_{t+1}$ during training. In this setup, frames are only used as input, and only the latent codes are subject to the CE loss. In this case, we retrain an IDM that only receives latent code inputs.
As shown in Tab.~\ref{table:supp_onlylc_go}, for both Go and robotic scenarios, using codes alone significantly improves performance, and incorporating next-frame prediction further enhances it. We hypothesize this is because next-frame prediction enhances the model's understanding of the environment and helps generate more accurate codes.
% \noindent \textbf{Data quality}
% 

% \vspace{-1mm}

\subsection{Visualizations}
We provide more visualizations of VideoWorld performing Go and robotic tasks in Fig.~\ref{fig:supp_vis_go} and Fig.~\ref{fig:supp_vis_calvin}, respectively. The Go visualizations are from matches between our 300M VideoWorld and KataGo-5d. With comparable Elo scores, these matches fairly demonstrate of how the model applies its learned knowledge. In Fig.~\ref{fig:supp_vis_cap}, our model captures opponent stones using the squeeze tactic. In Fig.~\ref{fig:supp_vis_sac}, VideoWorld deliberately sacrifices a single stone, prompting the opponent to capture it, thereby creating an opportunity to capture more of the opponent's stones. This highlights the model’s ability to prioritize long-term planning over short-term gains. For the robotic scenario, we show the model’s actual control results for the robotic arm, demonstrating its understanding of manipulation tasks and ability to execute them effectively.

\section{Conclusion}
\label{sec:discusion}

In this work, we take an initial step toward exploring knowledge learning from raw video data using the next token prediction paradigm. We conduct systematic experiments in custom Video Go and robotic simulation environments. We present two key findings: \textit{i}) The model can master the rules of Go and learn fundamental robotic operations, and \textit{ii}) The representation of visual change is crucial for knowledge learning. Based on these findings, we propose a latent dynamics model (LDM) to boost the efficiency and effectiveness of knowledge acquisition. Although applying this approach to real-world scenarios still faces challenges such as high-quality video generation and generalization, we believe that video generation models have the potential to serve as general knowledge learners and ultimately function as the artificial brain capable of thinking and acting in the real world.
\vspace{2em}

{
    \small
    \bibliographystyle{ieeenat_fullname}
    \bibliography{main}
}

% WARNING: do not forget to delete the supplementary pages from your submission 
\clearpage
\newpage
\appendix
\setcounter{page}{1}
\maketitlesupplementary

% \clearpage
% \setcounter{page}{1}
% \maketitlesupplementary
% \appendix

\renewcommand\thefigure{\thesection.\arabic{figure}}
\renewcommand\thetable{\thesection.\arabic{table}}
\setcounter{figure}{0} 
\setcounter{table}{0} 
\setcounter{table}{0}
\renewcommand{\thetable}{A.\arabic{table}}

% In this supplementary material, we first detail the training configuration, latent dynamics model, robotics action execution and inverse dynamics model in Sec.~\ref{sec:supp_detail}. Next, we present further analysis of latent dynamics model in Sec.~\ref{sec:supp_abla}. We also offer a comprehensive analysis of Video-GoBench in Sec.~\ref{sec:supp_dataset} and include more visualization results in Sec.~\ref{sec:supp_vis}.

% In this supplementary material, we first detail the training configurations, the latent dynamics model, robotics action execution and the inverse dynamics model in Sec.~\ref{sec:supp_detail}. Next, we present further analysis of the latent dynamics model in Sec.~\ref{sec:supp_abla}. We also offer a comprehensive analysis of Video-GoBench in Sec.~\ref{sec:supp_dataset} and include more visualization results in Sec.~\ref{sec:supp_vis}.

% In this supplementary material, we first detail the training configurations, the latent dynamics model, robotics action execution, and the inverse dynamics model in Sec.~\ref{sec:supp_detail}. Next, we present further analysis of the latent dynamics model in Sec.~\ref{sec:supp_abla}. We also offer a comprehensive analysis of Video-GoBench in Sec.~\ref{sec:supp_dataset} and include additional visualization results in Sec.~\ref{sec:supp_vis}.

\section{Implementation Details}
\label{sec:supp_detail}

\noindent \textbf{Training details.} We present the detailed training configurations of latent dynamics model and auto-regressive transformer in Tab.~\ref{table:supp_training_setting}.

\noindent \textbf{Latent dynamics model.}  The decoder can be divided into three parts based on three functions: \textit{i}) using an encoder to causally extract image features; \textit{ii}) using learnable embeddings to extract change information from the extracted features; \textit{iii}) using a decoder to causally reconstruct video frames based on image feature of the initial frame and the learned embeddings. We present the PyTorch-style pseudocodes for the overall LDM and each part in Alg.~\ref{alg:code}.

\noindent \textbf{Robotics action execution.}
Our model generates a set of latent codes $\{\hat{z}_t^h\}_{h=1}^H$ and next frame prediction $\hat{x}_{t+1}$ based on language instruction and input sequence $x_{1:t}$. We feed $\{\hat{z}_t^h\}_{h=1}^H$, $\hat{x}_{t+1}$ and $x_t$ into the Inverse Dynamics Model to obtain corresponding action. The inferred action is executed in an open-loop manner: after predicting each action, we input it into the environment engine to obtain a new observation $x_{t+1}$, which is then appended to the input sequence for the next prediction cycle. In Fig.~\ref{fig:umap_test_calvin}, we visualize the model's next-frame predictions and the actual control results of the robotic arm. We find that the model's next-frame predictions align with the task's execution intent and effectively control the robotic arm to complete the tasks. We visualize the latent codes at different time steps during testing and analyze them in Sec.~\ref{subsec:understand_ldm}.

\noindent \textbf{Inverse dynamics model.}
% The training objective of IDM is to directly predict the control signals for the robotic arm, a 7-dimensional vector that includes the arm's displacement along the XYZ axes, Euler angles, and the gripper's open/close state. Our IDM uses a ResNet-18 with an avgpool layer to process the mean squared error between two frames, resulting in a feature vector of size (1, 512). Simultaneously, IDM uses an MLP to process the latent codes' corresponding feature vectors into size ($H$, 512), $H$ is defined in Sec.3.2 and represents the total number of learnable embeddings in the LDM. Finally, another MLP with dimensions (512, 7) and an avgpool layer processes this ($H$+1, 512) vector into the final control signals.
% IDM is trained using the AdamW optimizer with a learning rate of 1e-4 for a total of 1M steps.
%The training objective of IDM is to directly predict the control signals for the robotic arm, which is a 7-dimensional vector including the arm's displacement along the XYZ axes, Euler angles, and the gripper's open/close state. Our IDM uses a ResNet-18 with an avgpool layer to process the frames to compute the mean squared error, resulting in a feature map and then avgpool it to size (1, 512). Simultaneously, the IDM uses an MLP to process the latent codes' corresponding feature vectors into a feature vector of size ($H$, 512). $H$ is defined in Section 3.2 and represents the total number of learnable embeddings in the LDM. Finally, another MLP with dimensions (512, 7) and an avgpool layer processes this vector of size ($H$+1, 512) into the final control signals.
The training objective of IDM is to predict the control signals for the robotic arm, represented as a 7-dimensional vector that includes the arm's displacement along the XYZ axes, Euler angles, and the gripper's open/close state. Specifically, IDM uses a ResNet-18 to process the predicted frame, applying a global average pooling layer to the penultimate feature map to obtain a feature vector of size (1, 512). Simultaneously, a shared MLP transforms the feature vectors of each latent code into a size of ($H$, 512), where $H$, as defined in Sec. 3.2, represents the number of learnable embeddings or prediction steps in LDM. Then, the features from the latent codes and the generated frames are concatenated into a vector of size ($H+1$, 512), which is then passed through another MLP with dimensions (512, 7) followed by a global average pooling layer to produce the final 7 control signals. IDM is trained using AdamW with a learning rate of 1e-4 for a total of 1 million steps, with mean squared error as the objective.

\noindent \textbf{RLBench evaluation.} In Sec.~\ref{subsec:res_ldm}, we test VideoWorld's ability to perform tasks in two different environments: CALVIN and RLBench. For CALVIN, we maintain the original task settings. For RLBench, we generate 20k trajectory data using a script. RLBench uses the same robotic arm and action space as CALVIN, but the environment appearance and task settings differ significantly. We use the combined RLBench and CALVIN datasets to train both the LDM and Transformer, while IDM is trained separately in each environment. Fig.~\ref{fig:supp_vis_rlbench} shows VideoWorld performing robotic tasks in RLBench environment.  

\begin{table}[t]
    \footnotesize
    \centering
    {
\begin{tabular}{p{2.8cm}|p{2.2cm} |p{2.2cm}}
    % \toprule
    % \multirow{2}*{Architecture} &\multicolumn{2}{c|}{MRSeg} &\multicolumn{2}{c}{refCOCOg} \\
     Config & LDM & AR Transformer\\

    \shline
    optimizer & AdamW &AdamW\\
     base learning rate & 5.4e-5 &3.0e-5\\
     weight decay & 0.01  &0\\
     optimizer monmentum & $\beta_1$, $\beta_2$=0.5,0.9 & $\beta_1$, $\beta_2$=0.9,0.98 \\
     batch size &16 & 256(Go),32(CALVIN) \\
     learning rate schedule & WarmipDecayLR & WarmipDecayLR \\
     warmup iterations & 3e+4 &3e+4\\
     max iterations &1e+5 &1e+6 \\
     augmentations & None & None \\
     Training Loss & L2 loss &Cross Entropy loss  \\
     Training Target & Reconstruction &Next token pred.  \\
     % $\lambda_{ref}$ & 2.0 \\
     % $\lambda_{dice}$ & 0.5 \\

\end{tabular}
}
\caption{\textbf{Training configurations} for the latent dynamics model (LDM) and auto-regressive (AR) transformer.}
\label{table:supp_training_setting}
\end{table}
% \input{tables/supp_baseline}
% \input{tables/supp_abla}
% \section{Extended Analysis and Ablative Studies}
% \label{sec:supp_abla}

% \begin{figure}[t]
% \includegraphics[width=\linewidth]{figures/umap_calvin.png}
% \centering
% % \vspace{-2em}
% \caption{\textbf{Illustration of robotic manipulation} and UMAP projection of the predicted latent code during inference. Latent codes are visualized through the LDM decoder. The UMAP projection illustrates the 9 predicted latent codes (i.e. $H=9$) across different tasks, with each point color-coded by task type. Visualizations with a yellow background show the model's actual robotic arm control during inference, while those with a green background represent the model's next-frame predictions during training.}
% \label{fig:umap_test_calvin}
% \vspace{-1.5em}
% \end{figure}

% \begin{figure*}[t]
%   % Shared Initialization
%   \begin{minipage}{0.5\textwidth}
    \begin{algorithm}[t]

      \caption{Pseudo codes of LDM.}
      \label{alg:code}
      \scriptsize
          \Comment{Inputs:~video:The first frame and its subsequent H frames [1+H, h, w, 3]; 
          }

          \Comment{Variables:~ldm\_q:~Learnable embeddings for H time spans [H,C]; image\_pe:position embedding of image features; 
         }
          \Comment{Functions:~CrossAttention();MLP();
          up\_scale(); down\_scale(); FSQ(); Causal3DCNN()}

    \Function{encoder(video)}{
    \Comment{video:video sequences.[1+H,h,w,3]}
    \Comment{The encoder consists of a set of encoder layers, composed of Causal3DCNN and down\_scale layers.}
    % \Comment{mask:output from the attention block above f.[Cx4Hx4W]}
    \var{f = Causal3DCNN(video)};\Comment{f:[1+H,h,w,C]}
    
    \For{layer in encoder\_layers}{
    \Comment{process and downsample features using Causal3DCNN and down\_scale.}
        \var{f = layer(f)}
    }

    \Comment{Capture dynamic changes in video. }
    \var{z = ldm\_qformer(f)}

    \var{return z, f[:0]};\Comment{f:[1+H,h',w',C]. z:[H,C]}
    
  }
  \Function{ldm\_qformer(f)}{
     \Comment{f:features of each frame.[1+H, h', w', C]}

    \var{q\_list = []} 
    
    \For{h in range(H)}{
        \var{query = ldm\_q[h]};\Comment{query:[1,C]}

        \var{f\_h = f[:(1+h)]};\Comment{f\_h:[1,h',w',C]}
        
        \var{key = f\_h + image\_pe};\Comment{key:[1,h',w',C]}

        \var{q\_h = CrossAttention(q=query, k=key, v=f\_h)};\Comment{q\_h:[1,C]}

        \var{q\_h = MLP(q\_h)};\Comment{q\_h:[1,C]}
        
        \var{q\_list.append(q\_h)}
    }

    \var{q\_list = stack(q\_list)};\Comment{q\_list:[H,C]}

    \var{return q\_list}
    
    }

  \Function{decoder(z, first\_f)}{
     \Comment{z:ldm query embeddings that captures change information in video frames.[H, C]; first\_f:image features of first frame.[1, h', w', C]}

    \Comment{
    The decoder consists of a set of decoder layers, composed of Causal3DCNN and up\_scale layers.
    }
    \var{z = repeat\_interleave(z, h', dim=-2)}

    \var{z = repeat\_interleave(z, w', dim=-1)}

    \var{rec\_video = cat(first\_f, z)}; \Comment{[1+H, h', w', C]}
    
    \For{layer in decoder\_layers}{
        \Comment{process and upsample features using Causal3DCNN and up\_scale.}
        \var{rec\_video = layer(rec\_video)}
    }

    \var{rec\_video = Causal3DCNN(rec\_video)} 
    
    \var{return rec\_video};\Comment{rec\_video:[1+H, h, w, C]}
  }

  \Comment{Main Function}
  \Function{latent\_dynamics\_model(video)}{
     \Comment{video:video sequences}

    \Comment{extract change information from video frames}
    \var{z, first\_f = encoder(video)}

    \var{z = FSQ(z)}; \Comment{quantize z using FSQ}
    
    \Comment{LDM training stage}
    \If{is\_train}{
        \Comment{obtain reconstructed video frames.}
        \var{rec\_video = decoder(z, first\_f)}

        \Comment{we train ldm using mse loss.}
       \var{ loss = MSE(rec\_video, video)}
        
        \var{return loss}
    }

    \var{return z}
    }

    \end{algorithm}

\begin{figure}[t]\centering
  \subfloat[\label{fig:state_count}Distribution of  board state count in training set]{\includegraphics[width=\linewidth]{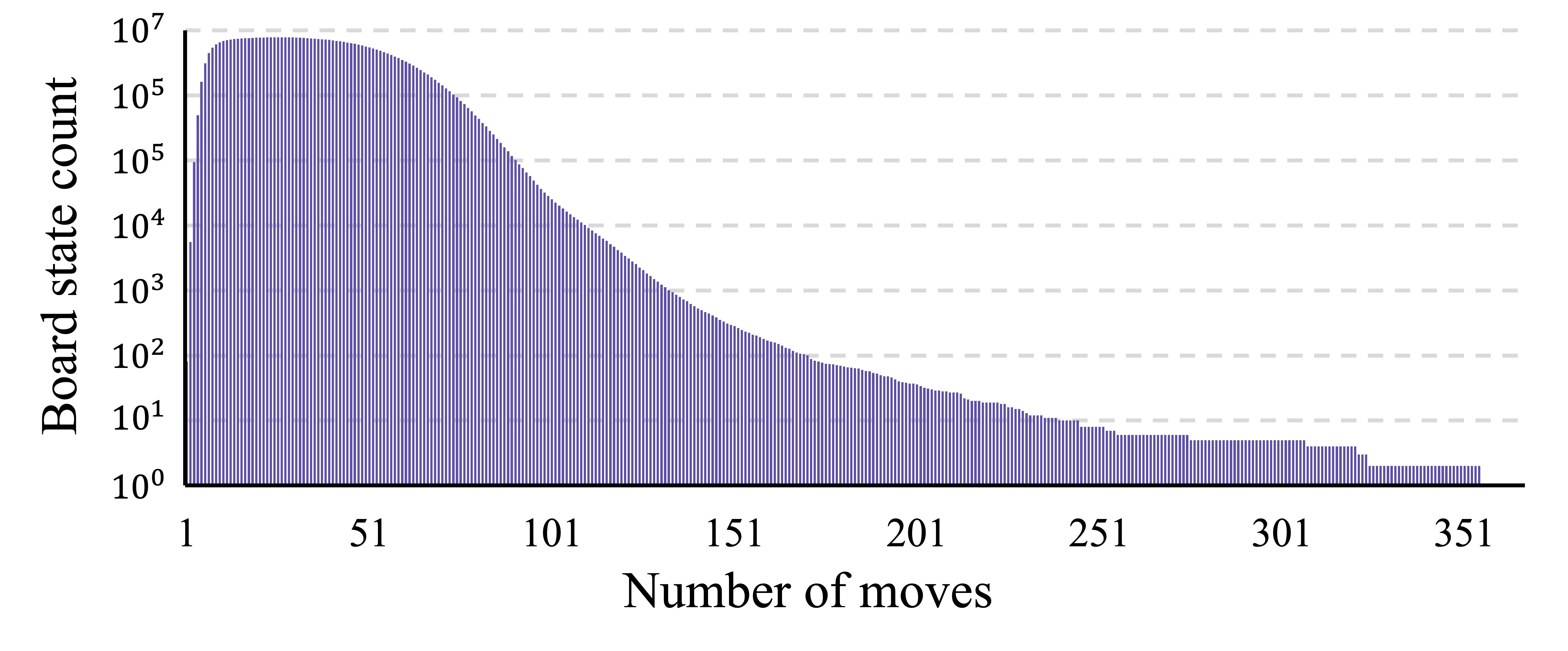}  }\\
% \hfill
\subfloat[\label{fig:state_rep}Repetition rate during online battle.]{\includegraphics[width=\linewidth]{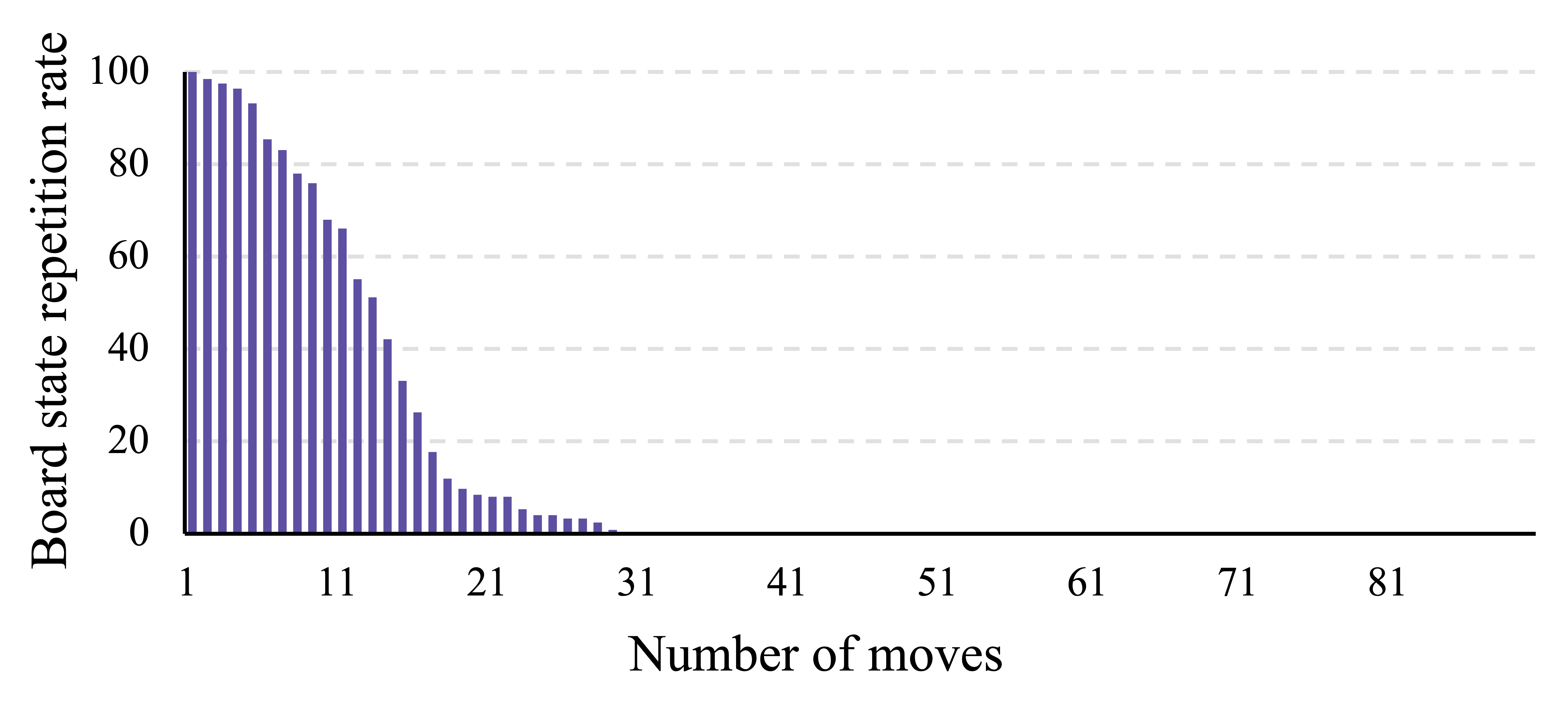} }
\hfill

\caption{\textbf{Dataset statistics}. Best viewed digitally.}
\vspace{-3mm}
\label{fig:supp_data}
\end{figure}

% \begin{figure}[t]\centering
%   \subfloat[\label{fig:supp_vis_cap}Capture opponent's stones]{\includegraphics[width=\linewidth]{figures/supp_vis_cap.png}  }\\
% % \hfil
% \vspace{-1em}
% \subfloat[\label{fig:supp_vis_sac}Sacrificing short-term gains to capture more opponent stones.]{\includegraphics[width=\linewidth]{figures/supp_vis_sac.png} }
% \hfill
% \vspace{-0.5em}
% \caption{\textbf{Visualizations of learned Go Strategies}. our model captures opponent stones using the squeeze tactic and self-sacrifice tactic. New black stones are red, new white stones are blue.}
% % \vspace{-1em}
% \label{fig:supp_vis_go}
% \end{figure}

% \begin{figure}[t]
% \includegraphics[width=\linewidth]{figures/supp_vis_calvin_3.png}
% \centering
% \vspace{-2em}
% \caption{\textbf{Visualizations of performing CALVIN tasks.}}
% \label{fig:supp_vis_calvin}
% \vspace{-1em}
% \end{figure}
% % Visualizations of VideoWorld performing robotic manipulation task} in CALVIN environment.

% \begin{figure}[t]
% \includegraphics[width=\linewidth]{figures/supp_vis_rlbench.png}
% \centering
% \vspace{-2em}
% \caption{\textbf{Visualizations of performing RLBench tasks.}}
% \label{fig:supp_vis_rlbench}
% \vspace{-1.5em}
% \end{figure}
% Visualizations of VideoWorld performing robotic manipulation task} in RLBench environment.

\section{Details on Video-GoBench}
\label{sec:supp_dataset}

% \subsection{Dataset Statistics}
In this section, we systematically analyze Video-GoBench. The benchmark includes many unique board states, offering extensive references for the model to learn from. Moreover, due to the combinatorial explosion effect, the proportion of repeated board states\textemdash those encountered during inference that also appear in the training set\textemdash decreases sharply as the game progresses when our model competes against reinforcement learning agents. This ensures that good performance cannot rely on memorizing training scenarios, highlighting the model’s generalization ability.

\noindent \textbf{Board state count.}
The Video-GoBench training set contains approximately 400M unique board states. We analyze their distribution based on move numbers, i.e., the number of stones each state contains. As shown in Fig.~\ref{fig:state_count}, the data features a diverse range of board states, primarily concentrated within the first 100 moves, with significantly fewer states beyond that.

\noindent \textbf{Board state repetition rate.}
We collect 400 game records between our model and KataGo-9d, calculating the overlap rate of board states with the training set across different move numbers. As shown in Fig.~\ref{fig:state_rep}, the repetition rate drops sharply as the games progress. By move 30, it reaches zero, with all games continuing beyond this point. This eliminates the possibility of relying on memory alone to achieve high performance.
% \begin{figure}[t]
% \includegraphics[width=\linewidth]{figures/supp_repetition_rate.png}
% \centering
% % \vspace{-2em}
% \caption{ }
% \label{fig:supp_repetition}
% \vspace{-1.5em}
% \end{figure}

% \section{Additional Visualizations}
% \label{sec:supp_vis}
% We provide more visualizations of VideoWorld performing Go and robotic tasks in Fig.~\ref{fig:supp_vis_go} and Fig.~\ref{fig:supp_vis_calvin}, respectively. The Go visualizations are from matches between our 300M VideoWorld and KataGo-5d. With comparable Elo scores, these matches fairly demonstrate of how the model applies its learned knowledge. In Fig.~\ref{fig:supp_vis_cap}, our model captures opponent stones using the squeeze tactic. In Fig.~\ref{fig:supp_vis_sac}, VideoWorld deliberately sacrifices a single stone, prompting the opponent to capture it, thereby creating an opportunity to capture more of the opponent's stones. This highlights the model’s ability to prioritize long-term planning over short-term gains. For the robotic scenario, we show the model’s actual control results for the robotic arm, demonstrating its understanding of manipulation tasks and ability to execute them effectively.
% {
%     \small
%     \bibliographystyle{ieeenat_fullname}
%     \bibliography{main}
% }

\end{document}